%% file: ms.tex
\documentclass[journal]{IEEEtran}
\pdfoutput=1

\IEEEoverridecommandlockouts                              

\usepackage[utf8]{inputenc}
\usepackage{graphicx}

\usepackage{amsmath} 
\usepackage{amsthm}  
\usepackage{amsfonts}
\usepackage{breqn}
\usepackage{dsfont}
\usepackage[dvipsnames]{xcolor}
\usepackage{tikz}
\usepackage[nolist,nohyperlinks]{acronym}
\usepackage[linesnumbered,ruled,vlined]{algorithm2e}
\usepackage[capitalize]{cleveref}
\Crefname{equation}{}{}
\usepackage[per-mode=symbol]{siunitx}
\usepackage{tikz}
\usepackage{pgfplots}
\pgfplotsset{compat=1.16}

\usepackage{url}
\usepackage[numbers,sort&compress]{natbib}
\usepackage{letltxmacro}
\LetLtxMacro{\autocite}{\cite}
\LetLtxMacro{\textcite}{\citet}

\theoremstyle{plain}
\newtheorem{definition}{Definition}
\theoremstyle{remark}\newtheorem{remarkenv}{Remark}        
\newenvironment{remark}{\begin{remarkenv}}%
	{\hfill$\lozenge$\end{remarkenv}}            
	
\usepackage{booktabs}
\usepackage{tabularx}
\newcommand{\otoprule}{\midrule[\heavyrulewidth]}	

\title{Scenario Parameter Generation Method and Scenario Representativeness Metric for Scenario-Based Assessment of Automated Vehicles}
\author{Erwin de Gelder$^{1,2*}$, Jasper Hof$^{3}$, Eric Cator$^{4}$, Jan-Pieter Paardekooper$^{1,5}$, Olaf Op den Camp$^{1}$, \\Jeroen Ploeg$^{6}$, Bart De Schutter$^{2}$%
\thanks{$^{1}$TNO, Integrated Vehicle Safety, Helmond, The Netherlands}%
\thanks{$^{2}$Delft University of Technology, Delft Center for Systems and Control, Delft, The Netherlands}%
\thanks{$^{3}$Radboud University Medical Center, Department for Health Evidence, Nijmegen, The Netherlands}%
\thanks{$^{4}$Radboud University, Applied Stochastics, Nijmegen, The Netherlands}%
\thanks{$^{5}$Radboud University, Donders Institute for Brain, Cognition and Behaviour, Nijmegen, The Netherlands}%
\thanks{$^{6}$Eindhoven University of Technology, Dept.\ of Mechanical Engineering Dynamics and Control group, Eindhoven, The Netherlands}%
\thanks{$^{*}$Corresponding author. \newline E-mail address: {\tt\small erwin.degelder@tno.nl}}}%
\date{}

\let\originalleft\left
\let\originalright\right
\renewcommand{\left}{\mathopen{}\mathclose\bgroup\originalleft}
\renewcommand{\right}{\aftergroup\egroup\originalright}
\newcommand{\bandwidth}{h}
  \newcommand{\bandwidthmatrix}{H}

\newcommand{\constraintmatrix}{A}
\newcommand{\constraintvector}{b}
\newcommand{\dirac}[1]{\delta\left(#1\right)}
\newcommand{\densitysymbol}{f}
  \newcommand{\density}[1]{\densitysymbol\left( #1 \right)}
  \newcommand{\densityestsymbol}{\hat{\densitysymbol}}
  \newcommand{\densityest}[1]{\densityestsymbol\left( #1 \right)}
  \newcommand{\densityestkde}[2]{\densityestsymbol_{#1}\left( #2 \right)}
  \newcommand{\densityissymbol}{g}
  \newcommand{\densityis}[1]{\densityissymbol\left(#1\right)}
\newcommand{\dimension}{d}
\newcommand{\distancefunc}[2]{\Delta\left(#1,#2\right)}
\newcommand{\dummyvar}{u}
  \newcommand{\dummyset}{\mathcal{U}}
  \newcommand{\dummyvarb}{v}
  \newcommand{\dummydensity}{\xi}
  \newcommand{\dummydensityb}{\eta}
  \newcommand{\dummydensityc}{\gamma}
  \newcommand{\dummydensityset}{\Gamma}
\newcommand{\e}[1]{\exp\left\{ #1 \right\}}
\newcommand{\elementindex}{k}
\newcommand{\elementproduct}{\odot}
  \newcommand{\elementdivision}{\oslash}
\newcommand{\extraparameters}{\theta}
  \newcommand{\dimensionextraparameters}{n_\theta}
\newcommand{\floor}[1]{\left\lfloor #1 \right\rfloor}
\newcommand{\funcmapping}[3]{#1: #2 \rightarrow #3}
\newcommand{\identitymatrix}[1]{I_{#1}}
\newcommand{\kernelfunc}[1]{K \left( #1 \right)}
  \newcommand{\kernelfuncnormalized}[2]{K_{#1} \left( #2 \right)}

\newcommand{\normtwo}[1]{\left\Vert #1 \right\Vert_2}
\newcommand{\penaltyweight}{\beta}
\newcommand{\proposedmetric}[4]{M_{#1}\left(#2,#3,#4\right)}
\newcommand{\realnumbers}{\mathds{R}}
\newcommand{\scenariopars}{\mathbf{x}}
  \newcommand{\scenarioparselement}[2]{\left(\scenariopars_{#1}\right)_{#2}}
  \newcommand{\scenarioparselementmean}[1]{\bar{\scenariopars}_{#1}}
  \newcommand{\dimensionscenariopars}{n_x}
  \newcommand{\scenariosnumberof}{N_x}
  \newcommand{\scenarioindex}{i}
  \newcommand{\scenarioindexb}{m}
  \newcommand{\scenarioparsmatrix}{X}
  \newcommand{\scenarioset}{\mathcal{X}}
  \newcommand{\scenarioparsreduced}{v}
  \newcommand{\scenarioparsgenerated}{\mathbf{w}}
  \newcommand{\scenariosgeneratednumberof}{N_w}
  \newcommand{\scenariosetgenerated}{\mathcal{W}}
  \newcommand{\scenarioparstest}{\mathbf{z}}
  \newcommand{\scenariostestnumberof}{N_z}
  \newcommand{\scenariosettest}{\mathcal{Z}}
  \newcommand{\scenariosetfake}{\scenarioset^*}
  \newcommand{\scenariosetgeneratedfake}{\scenariosetgenerated^*}
  \newcommand{\scenariosettestfakelarge}{\scenariosettest^*_{\mathrm{large}}}
  \newcommand{\scenariosettestfakesmall}{\scenariosettest^*}
\newcommand{\svdu}{U}
  \newcommand{\svduvec}[1]{\mathbf{u}_{#1}}
  \newcommand{\svds}{\Sigma}
  \newcommand{\svdsv}[1]{\sigma_{#1}}
  \newcommand{\svdv}{V}
  \newcommand{\svdventry}[2]{v_{#1 #2}}
  \newcommand{\svdvvecd}[1]{\tilde{\mathbf{v}}_{#1}}
  \newcommand{\svdrank}{\bar{N}}
  \newcommand{\svdindex}{j}
  \newcommand{\svdindexb}{k}
  \newcommand{\svdmean}{\mu}
\newcommand{\sumindex}{i}
  \newcommand{\sumindexb}{j}
\renewcommand{\time}{t}
  \newcommand{\timestart}{\time_0}
  \newcommand{\timeend}{\time_1}
\newcommand{\timeseries}[1]{y\left(#1\right)}
  \newcommand{\dimensiontimeseries}{n_y}
  \newcommand{\numberoftimesegments}{n_{\mathrm{t}}}
  \newcommand{\timeseriesvec}{\mathbf{y}}
\newcommand{\transpose}{^{\mkern-1.5mu\mathsf{T}}}
\newcommand{\transportmatrix}{T}
  \newcommand{\transportmatrixelement}[2]{T_{#1 #2}}
\newcommand{\ud}{\,\mathrm{d}}
\newcommand{\uniformvariable}{u}
  \newcommand{\uniformdist}[2]{U\left(#1,#2\right)}
\newcommand{\wassersteinemp}[3]{\tilde{W}_{#1}\left(#2,#3\right)}
  \newcommand{\wassersteinp}[3]{W_{#1}\left(#2,#3\right)}
  \newcommand{\wassersteincoefficient}{p}
\newcommand{\weights}{\alpha}
  \newcommand{\weight}[1]{\weights_{#1}}
  \newcommand{\weightpar}[1]{\beta_{#1}}

\newlength{\figurewidth}
\newlength{\figureheight}

\begin{document}

\begin{acronym}[AAAAAAAA]
	\acro{av}[AV]{Automated Vehicle}\acroindefinite{av}{an}{an}
	\acro{gan}[GAN]{Generative Adversarial Network}
	\acro{kde}[KDE]{Kernel Density Estimation}
	\acro{lvd}[LVD]{leading vehicle decelerating}\acroindefinite{lvd}{an}{a}
	\acro{pca}[PCA]{Principal Component Analysis}
	\acro{pdf}[pdf]{probability density function}
	\acro{sr}[SR]{Scenario Representativeness}\acroindefinite{sr}{an}{a}
	\acro{svd}[SVD]{Singular Value Decomposition}\acroindefinite{svd}{an}{a}
\end{acronym}

\maketitle

\input{secs/abstract}
\acresetall
\input{secs/introduction}
\input{secs/literature}
\input{secs/generation}
\input{secs/comparison}
\input{secs/example}
\input{secs/discussion}
\acresetall
\input{secs/conclusions}

\bibliographystyle{IEEEtranN}
\bibliography{bib}

\end{document}

%% file: secs/abstract.tex
\begin{abstract}

The development of assessment methods for the performance of \acp{av} is essential to enable the deployment of automated driving technologies, due to the complex operational domain of \acp{av}. 
One candidate is scenario-based assessment, in which test cases are derived from real-world road traffic scenarios obtained from driving data.
Because of the high variety of the possible scenarios, using only observed scenarios for the assessment is not sufficient. 
Therefore, methods for generating additional scenarios are necessary.

Our contribution is twofold. 
First, we propose a method to determine the parameters that describe the scenarios to a sufficient degree while relying less on strong assumptions on the parameters that characterize the scenarios. 
By estimating the \ac{pdf} of these parameters, realistic parameters values can be generated.
Second, we present the \ac{sr} metric based on the Wasserstein distance, which quantifies to what extent the scenarios with the generated parameter values are representative of real-world scenarios while covering the actual variety found in the real-world scenarios. 

A comparison of our proposed method with methods relying on assumptions of the scenario parameterization and \ac{pdf} estimation shows that the proposed method can automatically determine the optimal scenario parameterization and \ac{pdf} estimation.
Furthermore, it is demonstrated that our \ac{sr} metric can be used to choose the (number of) parameters that best describe a scenario.
The presented method is promising, because the parameterization and \ac{pdf} estimation can directly be applied to already available importance sampling strategies for accelerating the evaluation of \acp{av}.

\end{abstract}

%% file: secs/introduction.tex
\section{Introduction}
\label{sec:introduction}

\IEEEPARstart{A}{n} essential aspect in the development of \acp{av} is the assessment of the quality and performance of \ac{av} behavior with respect to safety, comfort, and efficiency \autocite{bengler2014threedecades, stellet2015taxonomy, koopman2016challenges}.
Because public road tests are expensive and time consuming \autocite{kalra2016driving, zhao2018evaluation}, a scenario-based approach has been proposed \autocite{riedmaier2020survey, elrofai2018scenario, putz2017pegasus, krajewski2018highD, deGelder2017assessment, stellet2015taxonomy, jacobo2019development}.
With a scenario-based approach, the response of the system-under-test is assessed in many scenarios and for the variations of these scenarios that occur in the real world.
Here, a scenario describes the situation the system-under-test is in and how this situation develops over time (in \cref{sec:parameterization}, a precise definition of the term scenario is provided).
One of the advantages of a scenario-based approach is that the assessment can focus on the more challenging situations by selecting scenarios that are challenging for the system-under-test.
As a source of information for the assessment scenarios, real-world driving data has been proposed, thereby guaranteeing that the scenarios represent real-world driving conditions \autocite{elrofai2018scenario, putz2017pegasus, krajewski2018highD}.

For the scenario-based assessment approach, it is important that the generated scenarios are representative of scenarios that could happen in real life. 
In other words, the scenarios should be a representation of the real world \autocite{riedmaier2020survey}.
Only then, the results of the assessment can be generalized to the performance of the system-under-test when operating in real life \autocite{deGelder2017assessment}.
Furthermore, it is essential that the generated scenarios cover the same variety that is found in real life. 
\textcite{riedmaier2020survey} argue that since an infinite number of situations occur in the real world, the scenario generation methods must provide a large number of variations in order to cover this infinite number of situations.

Our data-driven approach uses observed scenarios to generate parameter values that describe new scenarios.
Instead of relying on a predetermined functional form of the signals, such as a vehicle's speed, and fitting parameters to this functional form, we employ \iac{svd} \autocite{golub2013matrix} to determine in a data-driven manner the parameters that best describe the scenarios.
Next, the \ac{pdf} of the parameters is estimated, such that the \ac{pdf} can be used to sample the parameters to generate similar scenarios. 
To not assume a particular shape of the \ac{pdf}, 
\ac{kde} \autocite{rosenblatt1956remarks, parzen1962estimation} is used for the \ac{pdf} estimation. 
Furthermore, with \ac{kde}, the correlations that might exist among the parameters is modeled.
This work also proposes a novel metric called the \ac{sr} metric for quantifying to what extent the generated scenarios are representative and cover the actual variety of real-world scenarios. 
More specifically, this metric uses the Wasserstein distance \autocite{ruschendorf1985wasserstein} to compare a set of generated scenarios with a set of observed scenarios.

This article is organized as follows.
\Cref{sec:literature} reviews related works.
In \cref{sec:generation}, the approach for generating scenarios for the assessment of \acp{av} is explained. 
Next, \cref{sec:comparison} presents a novel metric for quantifying the performance of the scenario-generation method.
A case study is performed in \cref{sec:case study}.
\Cref{sec:discussion} discusses relevant implications of our approach and some directions for future research.
Conclusions of the paper are provided in \cref{sec:conclusions}.

%% file: secs/literature.tex
\section{Related works}
\label{sec:literature}

In this section, first, works concerning the generation of scenarios for the assessment of \acp{av} are reviewed.
Next, works related to the \ac{sr} metric are reviewed.

\subsection{Scenario generation}
\label{sec:literature generation}

The approaches to determine scenarios for the assessment of \acp{av} can be categorized into three kinds \autocite{li2016intelligence}: scenarios based on observations of real-world traffic, scenarios based on the functionality that is being assessed, and a combination of these two approaches.
The current paper focuses on the first approach.

In the literature, several methods are proposed to generate scenarios for the assessment based on real-world driving data. 
\textcite{lages2013automatic} proposed a method to construct scenarios in a virtual simulation environment by reconstructing the real-world scenarios observed by laser scanners.
\textcite{zofka2015datadrivetrafficscenarios} presented how recorded sensor data can be exploited to create scenarios that might lead to critical situations by modifying parameters of the recorded parameterized scenarios. 
\textcite{stepien2021applying} generate scenarios by sampling scenario parameter values from generalized extreme value distributions, where the distribution parameters are fitted using scenario parameter values extracted from safety-critical scenarios observed in naturalistic driving data.
In \autocite{feng2020testing, deGelder2017assessment, thal2020incorporating, feng2020safety, li2020theoretical, feng2021intelligent}, also parameterized scenarios were generated and, in addition, importance sampling techniques were presented that automatically generate scenarios in which the system-under-test shows (safety-)critical behavior. 
Other approaches to generate scenarios in which the system-under-test shows (safety-)critical behavior are Monte Carlo tree search \autocite{koren2018adaptive} and genetic programming \autocite{corso2020interpretable}.
\textcite{schuldt2018method} provided a method to generate scenarios using combinatorial algorithms that should ensure that the test cases cover the variety of the possible situations the system-under-test could encounter in real life.
More recently, \autocite{spooner2021generation} presented \iac{gan} to generate pedestrian crossing scenarios.

In the existing literature, the scenario generation methods for the assessment of \acp{av} have either one or more of the following shortcomings:
\begin{itemize}
	\item Observed scenarios are replayed without adding more variations \autocite{lages2013automatic}. 
	In this case, the total variety of scenarios that is found in real life will not be covered unless unrealistic amounts of data are gathered.
	
	\item The scenarios are oversimplified.
	For example, a vehicle's speed profile follows a predetermined functional form \autocite{deGelder2017assessment, thal2020incorporating, stepien2021applying}.
	
	\item Assumptions regarding the scenario parameter distributions are made that potentially compromise the quality of the scenarios.
	For example, the parameters are assumed to originate from a Gaussian \autocite{gietelink2007phd} or generalized extreme value \autocite{stepien2021applying} distribution, and/or it is assumed that (some of) the parameters are uncorrelated \autocite{feng2021intelligent}.
	
	\item Because no \ac{pdf} of the scenario parameters is known \autocite{zofka2015datadrivetrafficscenarios}, no evaluation can be made of the performance of the system once deployed on the road, because it is unknown how realistic and likely the scenarios are.
\end{itemize}
In \cref{sec:generation}, a method is proposed that overcomes these shortcomings.

\subsection{Scenario representativeness metric}
\label{sec:literature comparison}

The generated scenarios should represent scenarios that could happen in real life.
Whereas different approaches exist in the literature regarding the generation of scenarios for the assessment of \acp{av}, less is known about the comparison of the generated scenarios with real-life traffic.
From the mentioned sources in \cref{sec:literature generation}, only \textcite{feng2021intelligent} compared their generated scenarios with the ground truth from naturalistic driving data.
\textcite{feng2021intelligent} compared the distributions of vehicle speeds and bumper-to-bumper distances between the constructed scenarios and the ground truth.
To quantify the similarity between the distributions, the Hellinger distance \autocite{cha2007surveyPDFmeasures} and the mean absolute error were used.
The disadvantages of this approach are that:
\begin{enumerate}
	\item the generated scenarios may still be substantially different even though the distributions of the vehicle speeds and bumper-to-bumper distances are similar, and
	\item only the marginal distributions are considered while the correlation between the vehicle speeds and bumper-to-bumper distances might be completely different.
\end{enumerate}

Whereas little is known about comparing the generated scenarios for the scenario-based assessment of \acp{av} with ground truth data, many similarity metrics for comparing two \acp{pdf} are known \autocite{cha2007surveyPDFmeasures}.
Well-known metrics are the Minkowski metric \autocite{cha2007surveyPDFmeasures}, which is a generalized version of the Euclidean distance, the $f$-divergence, which is a generalized version of both the Kullback-Leibler divergence \autocite{kullback1951} and the Hellinger distance \autocite{cha2007surveyPDFmeasures}, and the Wasserstein metric \autocite{ruschendorf1985wasserstein}.
For practical reasons, this work uses the Wasserstein metric.
As is shown in \cref{sec:wasserstein}, the Wasserstein distance can be estimated using empirical distributions, i.e., without the need to estimate and evaluate \iac{pdf}.
The other mentioned metrics require integration over the domain of the \acp{pdf}, which will give computational issues since the considered \acp{pdf} will have a high dimensionality. 

%% file: secs/generation.tex
\section{Scenario generation}
\label{sec:generation}

To generate realistic scenarios for the assessment of \acp{av}, we use a data-driven approach: observed scenarios are used to generate new scenarios.
To do this, the scenarios are parameterized, i.e., parameters are defined that characterize a scenario. 
For example, the duration of a scenario could be a parameter. 
Next, the \ac{pdf} of the parameters is estimated.
This \ac{pdf} can be used to generate parameter values for new scenarios.
In addition, the \ac{pdf} contains the statistical information of the parameters so that the performance of \acp{av} can be estimated \autocite{deGelder2017assessment, zhao2016accelerated}. 
Choosing the parameters that describe a scenario, however, is not trivial:
\begin{itemize}
	\item Choosing too few parameters might lead to an oversimplification of the actual scenarios.
	As a result, not all possible variations of a scenario are modeled.
	
	\item Too many parameters lead to problems with estimating the \ac{pdf}, due to the curse of dimensionality \autocite{scott1992multivariate}.
\end{itemize}
To overcome this problem, we first consider as many parameters as needed for a complete description of the scenarios to avoid the oversimplification of the scenarios. 
Next, using \iac{svd}, a new set of parameters is created using a linear mapping of the original scenario parameters.
Because this new set of parameters is ordered according to the contribution of each of these parameters in describing the variation that exists among the original scenario parameters, we will consider only the most important parameters without losing too much information.
In this way, the curse of dimensionality is avoided without relying on a predetermined choice of parameters.

Below, we first explain how to describe a scenario using many parameters.
Next, \cref{sec:svd} proposes the use of the \ac{svd} to reduce the number of parameters.
\Cref{sec:kde} describes how \ac{kde} is used to estimate the \ac{pdf} of the reduced set of parameters and how the estimated \ac{kde} can be used to generate new scenarios.

\subsection{Parameterization of scenarios}
\label{sec:parameterization}

The first step of our approach is the parameterization of scenarios.
There is no single best way to parameterize the scenarios considering the wide variety of scenarios.
To deal with this variety, this work distinguishes quantitative scenarios from qualitative scenarios, using the definitions of \emph{scenario} and \emph{scenario category} of \autocite{degelder2020ontology}:

\begin{definition}[Scenario]
	\label{def:scenario}
	A scenario is a quantitative description of the relevant characteristics and activities and/or goals of the ego vehicle(s), the static environment, the dynamic environment and all events that are relevant to the ego vehicle(s) within the time interval between the first and last relevant event.
\end{definition}

\begin{definition}[Scenario category]
	\label{def:scenario category}
	A scenario category is a qualitative description of relevant characteristics and activities and/or goals of the ego vehicle(s), the static environment, and the dynamic environment.
\end{definition}

A scenario category is an abstraction of a scenario and, therefore, a scenario category comprises multiple scenarios \autocite{degelder2020ontology}.
For example, the scenario category ``cut-in'' comprises all possible cut-in scenarios.
The goal of our approach is to determine the optimal parameterization of scenarios of a given scenario category based on a set of observed scenarios of the same scenario category and to estimate the \ac{pdf} of these parameters that can be used to generate parameter values for new scenarios.

The observed scenarios are described using a time series for the content of the scenario that changes within the time window of the scenario (e.g., the speed of a vehicle) and some additional parameters for the content that is fixed (e.g., the lane width and the duration of the scenario).
Here, $\timeseries{\time} \in \realnumbers^{\dimensiontimeseries}$ denotes the time series of a scenario with $\time \in [\timestart, \timeend]$, where $\dimensiontimeseries$ denotes the dimension of the time series and $\timestart$ and $\timeend$ denote the start and end time of the scenario.
The $\dimensionextraparameters$ additional parameters are represented by $\extraparameters \in \realnumbers^{\dimensionextraparameters}$.

To deal with the time series, the continuous time interval $[\timestart, \timeend]$ is discretized, such that two consecutive time instants are $(\timeend - \timestart) / (\numberoftimesegments-1)$ apart. 
This gives:
\begin{equation}
	\label{eq:time series vectorized}
	\timeseriesvec = \begin{bmatrix}
		\timeseries{\timestart} \\ 
		\timeseries{\timestart + \frac{\timeend - \timestart}{\numberoftimesegments-1}} \\
		\timeseries{\timestart + 2\frac{\timeend - \timestart}{\numberoftimesegments-1}} \\
		\vdots \\
		\timeseries{\timeend}
	\end{bmatrix} \in \realnumbers^{\numberoftimesegments \dimensiontimeseries}.
\end{equation}
Note that $\numberoftimesegments$ must be chosen such that no important information is lost during the discretization.
Because in practice, due to the discrete nature of sensor readings, the time series $\timeseries{\time}$ is obtained at certain specific times rather than on a continuous time interval, it may be required to use interpolation techniques, such as splines \autocite{deboor1978practical}, to evaluate $\timeseriesvec$.

Let us assume that $\scenariosnumberof$ observed scenarios can be used to generate new scenarios. 
To indicate that the scenario parameters $\timeseriesvec$ and $\extraparameters$ belong to a specific scenario, the index $\scenarioindex \in \{1,\ldots,\scenariosnumberof\}$ is used, i.e., the parameters of the $\scenarioindex$-th scenario are $\timeseriesvec_{\scenarioindex}$ and $\extraparameters_{\scenarioindex}$.
To further ease the notation, $\timeseriesvec_{\scenarioindex}$ and $\extraparameters_{\scenarioindex}$ are combined into one vector $\scenariopars_{\scenarioindex}$:
\begin{equation}
	\label{eq:scenario parameters}
	\scenariopars_{\scenarioindex} = \begin{bmatrix}
		\timeseriesvec_{\scenarioindex} \\ \extraparameters_{\scenarioindex}
	\end{bmatrix} \in \realnumbers^{\numberoftimesegments\dimensiontimeseries + \dimensionextraparameters}.
\end{equation}

\subsection{Parameter reduction using \acl{svd}}
\label{sec:svd}

As shown in \cref{eq:scenario parameters}, $\dimensionscenariopars =  \numberoftimesegments\dimensiontimeseries + \dimensionextraparameters$ parameters describe a scenario.
Even for small numbers of $\numberoftimesegments$, $\dimensiontimeseries$, and $\dimensionextraparameters$, the total number of parameters becomes too large to reliably estimate the joint \ac{pdf}. 
One way to avoid this curse of dimensionality is to assume that the parameters are independent, but especially the parameters $\timeseries{\timestart}$ till $\timeseries{\timeend}$ in \cref{eq:time series vectorized} are obviously correlated, so assuming that the parameters are independent is not a good solution.

In the field of machine learning, \ac{pca} is commonly used for dimensionality reduction \autocite{abdi2010principal}.
As \ac{pca} uses the \ac{svd} \autocite{golub2013matrix}, this work uses the \ac{svd} to transform the parameters $\scenariopars_{\scenarioindex}$ into a lower-dimensional vector of parameters.
Before applying the \ac{svd}, the parameters are weighted with $\weights \in \realnumbers^{\dimensionscenariopars}$ in order to give more or less importance to the $\dimensionscenariopars$ parameters.
This is particularly useful to compensate for the imbalance in the parameter vector, where the imbalance is caused by the fact the parameter vector considers the time series $\timeseries{\time}$ at $\numberoftimesegments$ different times and the additional parameters $\extraparameters$ only once.
Let us define a matrix that contains the parameters of the $\scenariosnumberof$ scenarios:
\begin{equation}
	\label{eq:scenario parameter matrix}
	\scenarioparsmatrix = \begin{bmatrix}
		\left(\weights\elementproduct\scenariopars_1\right)-\svdmean & \cdots & \left(\weights\elementproduct\scenariopars_{\scenariosnumberof}\right)-\svdmean
	\end{bmatrix} \in \realnumbers^{\dimensionscenariopars \times \scenariosnumberof},
\end{equation}
where $\elementproduct$ denotes the element-wise product of vectors and $\svdmean\in\realnumbers^{\dimensionscenariopars}$ denotes the mean of the weighted scenario parameters:
\begin{equation}
	\svdmean = \frac{1}{\scenariosnumberof} \sum_{\scenarioindex=1}^{\scenariosnumberof} \weights \elementproduct\scenariopars_{\scenarioindex}.
\end{equation}

Using the \ac{svd} of $\scenarioparsmatrix$, we obtain:
\begin{equation}
	\label{eq:svd}
	\scenarioparsmatrix = \svdu \svds \svdv\transpose.
\end{equation}
Here, both $\svdu \in \realnumbers^{\dimensionscenariopars \times \dimensionscenariopars}$ and $\svdv \in \realnumbers^{\scenariosnumberof \times \scenariosnumberof}$ are orthonormal matrices.
Therefore, both matrices can be interpreted as rotation matrices in $\realnumbers^{\dimensionscenariopars}$ and $\realnumbers^{\scenariosnumberof}$, respectively.
The matrix $\svds \in \realnumbers^{\dimensionscenariopars \times \scenariosnumberof}$ takes the same shape as $\scenarioparsmatrix$.
This matrix has only zeros except on (part of) the diagonal.
The diagonal contains the so-called singular values, denoted by $\svdsv{\svdindex}$ with $\svdindex\in\{1,\ldots,\svdrank\}$, $\svdrank = \min(\dimensionscenariopars,\scenariosnumberof)$.
These singular values are in decreasing order, i.e.,
\begin{equation}
	\svdsv{1} \geq \svdsv{2} \geq \ldots \geq \svdsv{\svdrank} \geq 0.
\end{equation}
Because of the decreasing singular values, rotating the matrix $\scenarioparsmatrix$ from the left with $\svdu\transpose$ transforms the data to a new coordinate system such that the first coordinate has the largest variance compared to the other coordinates.
This variance equals $\svdsv{1}^2$.
Similarly, the second largest variance equals $\svdsv{2}^2$ and lies on the second coordinate, etc.
Because of the decreasing variance, the scenario parameters can be approximated using only the first $\dimension$ coordinates of the new coordinate system, as these $\dimension$ coordinates describe the majority of the variations.
So, the scenario parameters of the $\scenarioindex$-th scenario are approximated by setting $\svdsv{\svdindex}=0$ for $\svdindex > \dimension$:
\begin{equation}
	\label{eq:svd approximation}
	\weights \elementproduct \scenariopars_{\scenarioindex} 
	= \svdmean + \sum_{\svdindex=1}^{\svdrank} \svdsv{\svdindex} \svdventry{\scenarioindex}{\svdindex} \svduvec{\svdindex}
	\approx \svdmean + \sum_{\svdindex=1}^{\dimension} \svdsv{\svdindex} \svdventry{\scenarioindex}{\svdindex} \svduvec{\svdindex},
\end{equation}
where $\svdventry{\scenarioindex}{\svdindex}$ is the $(\scenarioindex,\svdindex)$-th element of $\svdv$, $\svduvec{\svdindex}$ is the $\svdindex$-th column of $\svdu$, and $\dimension$ is the number of parameters that are retained.
Thus, the $\dimensionscenariopars$ parameters of the $\scenarioindex$-th scenario are approximated using the $\dimension$ parameters $\svdventry{\scenarioindex}{1}, \ldots, \svdventry{\scenarioindex}{\dimension}$. 
The singular values $\svdsv{1}, \ldots, \svdsv{\dimension}$, the vectors $\svduvec{1}, \ldots, \svduvec{\dimension}$, and $\svdmean$ are used to map the new scenario parameters, $\svdventry{\scenarioindex}{1}, \ldots, \svdventry{\scenarioindex}{\dimension}$, to an approximation of the weighted original scenario parameters, $\weights\elementproduct\scenariopars_{\scenarioindex}$.

\begin{remark}
	Using the approximation of \cref{eq:svd approximation}, it is not necessary to evaluate the complete \ac{svd} of \cref{eq:svd}.
	Only the first $\dimension$ columns of $\svdu$ and $\svdv$ need to be computed and only the first $\dimension$ singular values. 
	In practice, $\dimension \ll \svdrank$, so this saves a significant amount of computation time.
\end{remark}

The choice of $\dimension < \svdrank$ is not trivial.
Choosing $\dimension$ too small results in too much loss of detail.
Choosing $\dimension$ too large will give problems when estimating the \ac{pdf} of the new parameters.
One method to choose $\dimension$ is to look at the amount of overall variance of $\alpha\elementproduct\scenariopars_{\scenarioindex}$ explained by the first $\dimension$ singular values.
The overall variance scales with the sum of the squared singular values \autocite[p.~77]{golub2013matrix}, i.e.,
\begin{equation}
	\label{eq:scaled variance}
	\sum_{\scenarioindex=1}^{\scenariosnumberof} \left( 
		\left(\weights\elementproduct\scenariopars_{\scenarioindex}\right)-\svdmean 
	\right)\transpose \left( 
		\left(\weights\elementproduct\scenariopars_{\scenarioindex}\right)-\svdmean 
	\right) 
	= \sum_{\svdindex=1}^{\svdrank} \svdsv{\svdindex}^2.
\end{equation}
Thus, the first $\dimension$ singular values explain
\begin{equation}
	\label{eq:explained variance}
	\frac{\sum_{\svdindex=1}^{\dimension} \svdsv{\svdindex}^2}{\sum_{\svdindex=1}^{\svdrank} \svdsv{\svdindex}^2}
\end{equation}
of the overall variance.
One approach would be to set $\dimension$ such that \cref{eq:explained variance} exceeds a certain threshold, such as 0.95.
Another way to choose $\dimension$ is by inspecting the actual approximation error in \cref{eq:svd approximation} and keep increasing $\dimension$ until the approximation error is not too large.
\Cref{sec:comparison} proposes an alternative way to determine $\dimension$ using a metric that quantifies the goal of our generated scenarios, i.e., that the generated scenarios are representing real-world scenarios and cover the actual variety of real-world scenarios.

\subsection{Estimating the \acl{pdf}}
\label{sec:kde}

Using the approximation of \cref{eq:svd approximation} based on the \ac{svd}, the $\scenarioindex$-th scenario is described by the vector $\svdvvecd{\scenarioindex}$:
\begin{equation}
	\label{eq:reduced parameter vector}
	\svdvvecd{\scenarioindex}\transpose = \begin{bmatrix}
		\svdventry{\scenarioindex}{1} & \cdots & \svdventry{\scenarioindex}{\dimension}
	\end{bmatrix}.
\end{equation}
Note that the $\dimension$ entries of $\svdvvecd{\scenarioindex}$ are linearly uncorrelated with the $\dimension$ entries of $\svdvvecd{\scenarioindexb}$ ($\scenarioindexb\ne\scenarioindex$)\footnote{ 
	This is assuming that $\svdsv{\dimension}>0$. With this assumption and because $\scenarioparsmatrix$ in \cref{eq:scenario parameter matrix} is defined such that the sum of each row of $\scenarioparsmatrix$ equals zero, it is easy to verify that $\frac{1}{\scenariosnumberof}\sum_{\scenarioindexb=1}^{\scenariosnumberof} \svdventry{\scenarioindexb}{\svdindex}=0$ for $\svdindex\in\{1,\ldots,\scenariosnumberof\}$.  
	Therefore, $\sum_{\scenarioindex=1}^{\scenariosnumberof} \left( \svdventry{\scenarioindex}{\svdindex} - \frac{1}{\scenariosnumberof}\sum_{\scenarioindexb=1}^{\scenariosnumberof} \svdventry{\scenarioindexb}{\svdindex} \right)\left( \svdventry{\scenarioindex}{\svdindexb} - \frac{1}{\scenariosnumberof}\sum_{\scenarioindexb=1}^{\scenariosnumberof} \svdventry{\scenarioindexb}{\svdindexb} \right)=\sum_{\scenarioindex=1}^{\scenariosnumberof} \svdventry{\scenarioindex}{\svdindex}\svdventry{\scenarioindex}{\svdindexb}=0$ for $\svdindex\ne\svdindexb$, where the latter equality follows from the orthonormality of $\svdv$.}.
Despite the linear independence, the different entries of $\svdvvecd{\scenarioindex}$ may still be dependent due to higher-order correlations; so we treat these $\dimension$ entries as dependent variables.

To estimate the \ac{pdf} of $\svdvvecd{\scenarioindex}$, we propose to use \ac{kde}.
\ac{kde} \autocite{rosenblatt1956remarks, parzen1962estimation} is often referred to as a non-parametric way to estimate the \ac{pdf}, because \ac{kde} does not rely on the assumption that the data are drawn from a given parametric family of probability distributions.
Because \ac{kde} produces \iac{pdf} that adapts itself to the data, it is flexible regarding the shape of the actual underlying distribution of $\svdvvecd{\scenarioindex}$.
In \ac{kde}, the \ac{pdf} is estimated as:
\begin{equation}
	\label{eq:density est kde}
	\densityestkde{\bandwidthmatrix}{\scenarioparsreduced} = \frac{1}{\scenariosnumberof} 
	\sum_{\scenarioindex=1}^{\scenariosnumberof} \kernelfuncnormalized{\bandwidthmatrix}{\scenarioparsreduced - \svdvvecd{\scenarioindex}}.
\end{equation}
Here, $\kernelfuncnormalized{\bandwidthmatrix}{\cdot}$ is the so-called scaled kernel with a positive definite symmetric bandwidth matrix $\bandwidthmatrix \in \realnumbers^{\dimension \times \dimension}$.
The kernel $\kernelfunc{\cdot}$ and the scaled kernel $\kernelfuncnormalized{\bandwidthmatrix}{\cdot}$ are related using
\begin{equation}
	\kernelfuncnormalized{\bandwidthmatrix}{\dummyvar} = \left|\bandwidthmatrix\right|^{-1/2} \kernelfunc{\bandwidthmatrix^{-1/2} \dummyvar},
\end{equation}
where $|\cdot|$ denotes the matrix determinant. 
The choice of the kernel function is not as important as the choice of the bandwidth matrix \autocite{turlach1993bandwidthselection, duong2007ks}.
This article considers the Gaussian kernel\footnote{The advantage of the Gaussian kernel is that it gives the possibility to calculate a metric that quantifies the completeness of the data \autocite{degelder2019completeness} and to apply conditional sampling when generating scenario parameters \autocite{degelder2021conditional}. Both these topics are out of scope of this article.}, which is given by
\begin{equation}
	\label{eq:gaussian kernel}
	\kernelfunc{\dummyvar} = \frac{1}{\left(2\pi\right)^{\dimension/2}} \e{-\frac{1}{2} \normtwo{\dummyvar}^2},
\end{equation}
where $\normtwo{\dummyvar}^2=\dummyvar\transpose \dummyvar$ denotes the squared 2-norm of $\dummyvar$.

A bandwidth matrix of the form $\bandwidthmatrix=\bandwidth^2 \identitymatrix{\dimension}$ is used, where $\identitymatrix{\dimension}$ denotes the $\dimension$-by-$\dimension$ identity matrix.
The bandwidth $\bandwidth$ is determined with leave-one-out cross-validation \autocite{duin1976parzen}, because this minimizes the difference between the real \ac{pdf} and the estimated \ac{pdf} according to the Kullback-Leibler divergence \autocite{turlach1993bandwidthselection, zambom2013review}.

To sample scenario parameters using $\densityestkde{\bandwidthmatrix}{\cdot}$, first, an integer $\scenarioindex \in \{1,\ldots,\scenariosnumberof\}$ is randomly chosen with each integer having equal likelihood. 
Next, a random sample is drawn from a Gaussian with covariance $\bandwidthmatrix$ and mean $\svdvvecd{\scenarioindex}$.
Then, using the approximation in \cref{eq:svd approximation}, the scenario parameters are calculated.

As far as the computational effort is concerned, sampling the scenario parameters from \iac{kde} is efficient because there is no need to actually evaluate the \ac{pdf}. 
Determining the optimal bandwidth matrix requires more computational effort, but this only has to be done once per data set. 
The computational complexity of cross-validation methods for the bandwidth estimation typically scales with $\scenariosnumberof^2$ \autocite{gramacki2018nonparametric}.

%% file: secs/comparison.tex
\section{Scenario Representativeness metric}
\label{sec:comparison}

Ideally, the parameters of the generated scenarios are sampled from the same distribution that underlies the real-world scenario parameters. 
The problem is that this distribution is unknown. 
Nevertheless, it is possible to define a metric that quantifies the similarity of the distribution that is used to generate scenario parameters  and the distribution that underlies the real-world scenario parameters.
\Cref{sec:comparion problem} further explains the goal of this metric, which we call the \ac{sr} metric.
Next, \cref{sec:wasserstein} explains the Wasserstein distance \autocite{ruschendorf1985wasserstein}, which is then applied to derive our metric in \cref{sec:metric scenario generation method}.

\subsection{Scenario comparison problem}
\label{sec:comparion problem}

The set of observed scenarios, described using the parameters $\scenariopars_{\scenarioindex}$, $\scenarioindex \in \{1,\ldots,\scenariosnumberof\}$, are used for generating the scenario parameters. 
To ease the notation, let us denote the set of observed scenarios by $\scenarioset = \{\scenariopars_{1}, \ldots, \scenariopars_{\scenariosnumberof}\}$.
This work assumes that these scenarios --- that are comprised by the same scenario category --- are independently and identically distributed according to the distribution $\funcmapping{\density{\cdot}}{\realnumbers^{\dimensionscenariopars}}{\realnumbers}$.
Let us denote the set of generated scenario parameter vectors by $\scenariosetgenerated = \{\scenarioparsgenerated_{1}, \ldots, \scenarioparsgenerated_{\scenariosgeneratednumberof}\}$ where $\scenarioparsgenerated_{\scenarioindex} \in \realnumbers^{\dimensionscenariopars}$, $\scenarioindex \in \{1,\ldots,\scenariosgeneratednumberof\}$ are similarly parameterized as in \cref{eq:scenario parameters} and $\scenariosgeneratednumberof$ is the number of generated scenario parameter vectors.
Let $\funcmapping{\densityest{\cdot}}{\realnumbers^{\dimensionscenariopars}}{\realnumbers}$ denote the \ac{pdf} of the generated scenario parameter vectors, which is obtained from 
$\funcmapping{\densityestkde{\bandwidthmatrix}{\cdot}}{\realnumbers^{\dimension}}{\realnumbers}$ under a change of variable according to the approximation in \cref{eq:svd approximation}.
As later appears, it is not needed to have an explicit definition for $\densityest{\cdot}$.
Ideally, $\densityest{\cdot}$ is equal to $\density{\cdot}$.
So our metric aims to quantify the similarity of $\densityest{\cdot}$ and $\density{\cdot}$.

To estimate the similarity between $\densityest{\cdot}$ and $\density{\cdot}$, we cannot simply compare $\scenariosetgenerated$ with $\scenarioset$.
In that case, taking $\scenariosetgenerated=\scenarioset$ would give us the best result, but this is undesirable because, ideally, the scenarios of the generated parameters cover the whole variety of real-world scenarios and not just the variety that have been observed in $\scenarioset$.
Therefore, another set of scenarios is needed that can be used to test.
Let us assume that such a set of scenarios is available, denoted by $\scenariosettest = \{\scenarioparstest_{1},\ldots,\scenarioparstest_{\scenariostestnumberof}\}$ where $\scenarioparstest_{\scenarioindex} \in \realnumbers^{\dimensionscenariopars}$, $\scenarioindex \in \{1, \ldots, \scenariostestnumberof\}$ are independently and identically distributed according to $\density{\cdot}$.
Thus, $\scenarioset$ and $\scenariosettest$ can be regarded as a training and test set, respectively.

In summary, the goal is to find a metric that quantifies the similarity of $\densityest{\cdot}$ and $\density{\cdot}$ using the sets of observed scenario parameters $\scenarioset$ and $\scenariosettest$ and the set of scenario parameters $\scenariosetgenerated$, generated based on $\scenarioset$.

\subsection{Empirical Wasserstein metric}
\label{sec:wasserstein}

The $\wassersteincoefficient$-th Wasserstein metric ($\wassersteincoefficient \geq 1$) \autocite{ruschendorf1985wasserstein} is used to compare two \acp{pdf} $\dummydensity(\cdot)$ and $\dummydensityb(\cdot)$ defined on the set $\dummyset$.
This metric is defined as follows:
\begin{equation}
	\label{eq:wasserstein}
	\wassersteinp{\wassersteincoefficient}{\dummydensity}{\dummydensityb}
	= \left( 
		\inf_{\dummydensityc \in \dummydensityset(\dummydensity,\dummydensityb)}
		\left\{ 
			\int_{\dummyset\times\dummyset}
			\left( \distancefunc{\dummyvar}{\dummyvarb} \right)^{\wassersteincoefficient} 
			\ud \dummydensityc(\dummyvar,\dummyvarb)
		\right\}
	\right)^{1/\wassersteincoefficient}.
\end{equation}
Here, $\distancefunc{\dummyvar}{\dummyvarb}$ denotes the distance from $\dummyvar$ to $\dummyvarb$, which will be defined below, and $\dummydensityset(\dummydensity,\dummydensityb)$ denotes the set of joint distributions of $(\dummyvar, \dummyvarb)$ that have marginal distributions $\dummydensity(\cdot)$ and $\dummydensityb(\cdot)$. 
Intuitively, if the \acp{pdf} $\dummydensity(\cdot)$ and $\dummydensityb(\cdot)$ are seen as two piles of earth having a different shape with mass 1, then \cref{eq:wasserstein} calculates the minimum cost of converting one pile of earth with shape $\dummydensity(\cdot)$ into a pile of earth with shape $\dummydensityb(\cdot)$. 
Therefore, the Wasserstein metric is also referred to as the earth mover's distance \autocite{rubner2000emd}.

In our case, the goal is to have a metric to compare $\density{\cdot}$ and $\densityest{\cdot}$.
Because $\density{\cdot}$ is unknown, its approximation based on $\scenariosettest$ is considered:
\begin{equation}
	\label{eq:empirical density}
	\density{\scenarioparstest} \approx
	\frac{1}{\scenariostestnumberof} \sum_{\scenarioindex=1}^{\scenariostestnumberof}
	\dirac{\scenarioparstest - \scenarioparstest_{\scenarioindex}},
	\scenarioparstest\in\realnumbers^{\dimensionscenariopars},
\end{equation}
where $\dirac{\cdot}$ denotes the Dirac delta function.
Considering the high dimension of $\scenarioparstest$, numerical approximation of the integral of the  Wasserstein metric (14) using this approximation and $\densityest{\cdot}$ would require so many evaluations of $\densityest{\cdot}$ that it becomes computationally infeasible.
Therefore, the empirical estimation of the Wasserstein metric (14) is considered, which makes use of the empirical estimation of $\densityest{\cdot}$:
\begin{equation}
	\label{eq:empirical density estimation}	
	\densityest{\scenarioparsgenerated} \approx
	\frac{1}{\scenariosgeneratednumberof} \sum_{\scenarioindex=1}^{\scenariosgeneratednumberof}
	\dirac{\scenarioparsgenerated - \scenarioparsgenerated_{\scenarioindex}},
	\scenarioparsgenerated\in\realnumbers^{\dimensionscenariopars}.
\end{equation}
Substituting the empirical estimations of \cref{eq:empirical density,eq:empirical density estimation} for $\dummydensity(\cdot)$ and $\dummydensityb(\cdot)$, respectively, into \cref{eq:wasserstein}, leads to the so-called empirical Wasserstein metric \autocite{sommerfeld2018inference}, which is defined as:
\begin{equation}
	\label{eq:empirical wasserstein}
	\wassersteinemp{\wassersteincoefficient}{\scenariosettest}{\scenariosetgenerated}
	= \left( \inf_{\transportmatrix} 
	\sum_{\sumindex=1}^{\scenariostestnumberof} \sum_{\sumindexb=1}^{\scenariosgeneratednumberof} 
	\left(\distancefunc{\scenarioparstest_{\sumindex}}{\scenarioparsgenerated_{\sumindexb}}\right)^{\wassersteincoefficient}
	\transportmatrixelement{\sumindex}{\sumindexb}\right)^{1/\wassersteincoefficient},
\end{equation}
where $\transportmatrixelement{\sumindex}{\sumindexb}$ is the $(\sumindex,\sumindexb)$-th element of the transportation matrix $\transportmatrix$ that is subject to the following conditions:
\begin{alignat}{2}
	\sum_{\sumindex=1}^{\scenariostestnumberof} \transportmatrixelement{\sumindex}{\sumindexb} &= \frac{1}{\scenariosgeneratednumberof} \quad && \forall\ \sumindexb \in \{1,\ldots,\scenariosgeneratednumberof\}, \\
	\sum_{\sumindexb=1}^{\scenariosgeneratednumberof} \transportmatrixelement{\sumindex}{\sumindexb} &= \frac{1}{\scenariostestnumberof} \quad && \forall\ \sumindex \in \{1,\ldots,\scenariostestnumberof\}, \\
	\transportmatrixelement{\sumindex}{\sumindexb} &\geq 0 \quad && \forall\ \sumindex \in \{1,\ldots,\scenariostestnumberof\}, \sumindexb \in \{1,\ldots,\scenariosgeneratednumberof\}
\end{alignat}
For the distance function, we will use the 2-norm of the difference of the scenario parameters after scaling the scenario parameters according to the weights $\weights$ that we also used in \cref{sec:svd}:
\begin{equation}
	\distancefunc{\scenarioparstest}{\scenarioparsgenerated} = \normtwo{\left(\weights\elementproduct\scenarioparstest\right) - \left(\weights\elementproduct\scenarioparsgenerated\right)}.
\end{equation}

\subsection{Metric for testing scenario representativeness}
\label{sec:metric scenario generation method}

The empirical Wasserstein metric $\wassersteinemp{\wassersteincoefficient}{\scenariosettest}{\scenariosetgenerated}$ is an approximation of the Wasserstein metric $\wassersteinp{\wassersteincoefficient}{\densitysymbol}{\densityestsymbol}$.
As one might expect, using an infinite number of scenario parameters, i.e., for $\scenariostestnumberof\rightarrow\infty$ and $\scenariosgeneratednumberof\rightarrow\infty$, the empirical Wasserstein metric approaches the Wasserstein metric with probability 1 \autocite{sommerfeld2018inference}.
The problem is that $\scenariostestnumberof$ and $\scenariosgeneratednumberof$ are not infinite. 
In addition, whereas a fairly large number for $\scenariosgeneratednumberof$ can be chosen, as it is only limited by the available computational resources, to increase $\scenariostestnumberof$, more data are needed and this is generally expensive. 
Therefore, this work proposes a metric that is different from \cref{eq:empirical wasserstein}.

Our proposed \ac{sr} metric is based on the following intuition: Suppose that $\densityestsymbol$ is indeed an approximation of $\densitysymbol$. 
Because $\scenarioset$ and $\scenariosettest$ are based on the same underlying \ac{pdf}, i.e., $\densitysymbol$, it is expected that $\wassersteinemp{\wassersteincoefficient}{\scenarioset}{\scenariosetgenerated}$ is similar to $\wassersteinemp{\wassersteincoefficient}{\scenariosettest}{\scenariosetgenerated}$.
If, however, $\wassersteinemp{\wassersteincoefficient}{\scenarioset}{\scenariosetgenerated}$ is significantly smaller than $\wassersteinemp{\wassersteincoefficient}{\scenariosettest}{\scenariosetgenerated}$, it suggests overfitting of the training data because the generated scenario parameters are too much skewed towards the training data $\scenarioset$.
To penalize overfitting of the training data, our \ac{sr} metric includes a penalty in case $\wassersteinemp{\wassersteincoefficient}{\scenariosettest}{\scenariosetgenerated}$ is larger than $\wassersteinemp{\wassersteincoefficient}{\scenarioset}{\scenariosetgenerated}$.
Thus, the \ac{sr} metric becomes:
\begin{dmath}
	\label{eq:proposed metric}
	\proposedmetric{\wassersteincoefficient}{\scenariosetgenerated}{\scenariosettest}{\scenarioset}
	= \wassersteinemp{\wassersteincoefficient}{\scenariosettest}{\scenariosetgenerated} + \\
	\penaltyweight\left( \wassersteinemp{\wassersteincoefficient}{\scenariosettest}{\scenariosetgenerated} - \wassersteinemp{\wassersteincoefficient}{\scenarioset}{\scenariosetgenerated} \right).
\end{dmath}
Here, $\penaltyweight$ is the weight of the penalty. 
The case study in \cref{sec:case study} demonstrates empirically that $\proposedmetric{\wassersteincoefficient}{\scenariosetgenerated}{\scenariosettest}{\scenarioset}$ of \cref{eq:proposed metric} better correlates with the Wasserstein metric of \cref{eq:wasserstein} than the empirical Wasserstein metric of \cref{eq:empirical wasserstein} and a method to choose $\penaltyweight$.

%% file: secs/example.tex
\section{Case study}
\label{sec:case study}

To illustrate the proposed method for generating the scenario parameters (\cref{sec:generation}) and the \ac{sr} metric (\cref{sec:comparison}), these are applied in a case study.
\Cref{sec:case study intro} explains the scenario categories that are considered in the case study and describes the choices that are made regarding the scenario parameterization.
Next, the scenario parameter generation method is demonstrated in \cref{sec:case study parameter reduction}. 
\Cref{sec:case study parameter reduction} also shows that the \ac{sr} metric \cref{eq:proposed metric} can be used to choose $\dimension$.
Our method for generating scenario parameters is compared with other methods in \cref{sec:case study comparison method}.
\Cref{sec:case study comparison metric} demonstrates that the \ac{sr} metric \cref{eq:proposed metric} better correlates with the Wasserstein metric \cref{eq:wasserstein} than the empirical Wasserstein metric \cref{eq:empirical wasserstein}.

\subsection{Scenario categories and parameterization}
\label{sec:case study intro}

\begin{figure}
	\centering
	\input{figs/lead_braking.tikz}
	\caption{Schematic representation of the scenario category ``\acf{lvd}''. 
		The left vehicle is the ego vehicle.}
	\label{fig:lead vehicle decelerating}
\end{figure}
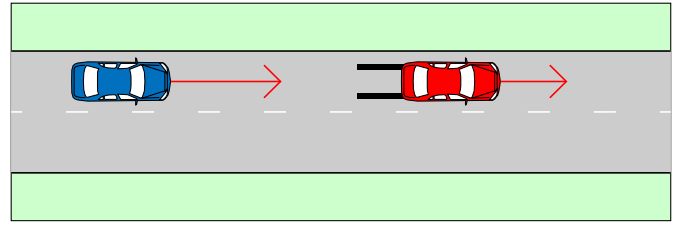

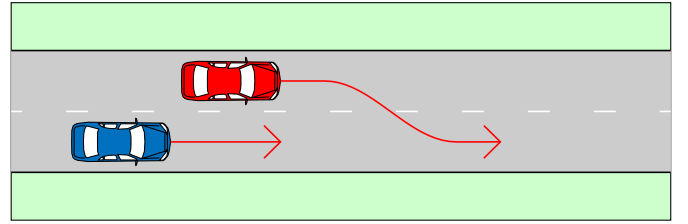
\begin{figure}
	\centering
	\input{figs/cut-in.tikz}
	\caption{Schematic representation of the scenario category ``cut-in''. 
		The left vehicle is the ego vehicle.}
	\label{fig:cut in}
\end{figure}

In this case study, two scenario categories are considered.
The first scenario category, labeled \ac{lvd}, involves an ego vehicle that is following another vehicle that decelerates, see \cref{fig:lead vehicle decelerating}.
As a result, the ego vehicle might need to brake or change direction to avoid contact with the vehicle that decelerates.
The second scenario category considers a vehicle that performs a cut-in, such that this vehicle becomes the leading vehicle of the ego vehicle, see \cref{fig:cut in}.
Depending on the speed and timing of the vehicle that performs a cut-in, the ego vehicle might need to brake or change direction to avoid a collision.

\setlength{\figurewidth}{.99\linewidth}

To obtain the scenarios, the data set described in \autocite{paardekooper2019dataset6000km} is used. 
The data were recorded from a single vehicle in which 20 drivers were asked to drive a prescribed route, resulting in 63 hours of data containing 1150 \ac{lvd} scenarios and 289 cut-in scenarios.
The majority of the route was on the highway.
To measure the surrounding traffic, the vehicle was equipped with three radars and one camera. 
The surrounding traffic was measured by fusing the data of the radars and the camera as described in  \autocite{elfring2016effective}.
To extract the \ac{lvd} and cut-in scenarios from the data set with the fused data, we searched for particular (combination) of activities in the data: a deceleration activity of a leading vehicle indicates \iac{lvd} scenario and a lane change of another vehicle that becomes the leading vehicle indicates a cut-in scenario. 
For more information on the process of extracting the scenarios, see \autocite{degelder2020scenariomining}.

\begin{figure}
	\centering
	\input{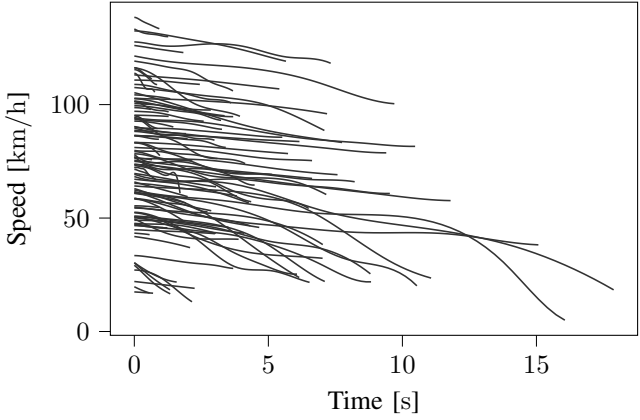}
	\caption{Speed of the leading vehicle during 100 randomly-selected observed \ac{lvd} scenarios. 
		For plotting purposes, the starting time of each scenario is set to 0.}
	\label{fig:speed lvd observed}
\end{figure}

From the 1150 \ac{lvd} scenarios, the training uses \SI{80}{\percent} (so $\scenariosnumberof=920$) and the testing uses the remaining \SI{20}{\percent} (so $\scenariostestnumberof=230$) as this 80/20 ratio is commonly used for splitting the data into a training set and a test set.
The training data are used for generating $\scenariosgeneratednumberof=10000$ new scenario parameter vectors.
To describe the decelerating behavior of the leading vehicle, the acceleration of the leading vehicle at $\numberoftimesegments=50$ time instants is used ($\dimensiontimeseries=1$). 
As additional parameters, the duration of the scenario, $\timeend-\timestart$, the initial speed of the leading vehicle, and the initial time gap between the leading vehicle and the ego vehicle are considered ($\dimensionextraparameters=3$). 
Thus, $\dimensionscenariopars=53$.
In \cref{fig:speed lvd observed}, the speed of the leading vehicle of 100 randomly-selected observed \ac{lvd} scenarios are shown.
The $\elementindex$-th weight, $\weight{\elementindex}$, is obtained by dividing a chosen constant $\weightpar{\elementindex}$ by the standard deviation of the $\elementindex$-th parameter:
\begin{equation}
	\label{eq:weights calculation}
	\weight{\elementindex} = \frac{\weightpar{\elementindex}}{\sqrt{
		\frac{1}{\scenariosnumberof}
		\sum_{\scenarioindex=1}^{\scenariosnumberof}
		\left( \scenarioparselement{\scenarioindex}{\elementindex} - \scenarioparselementmean{\elementindex} \right)^2
	}},
\end{equation} 
with $\scenarioparselement{\scenarioindex}{\elementindex}$ denoting the $\elementindex$-th element of $\scenariopars_{\scenarioindex}$ and $\scenarioparselementmean{\elementindex}=\frac{1}{\scenariosnumberof}\sum_{\scenarioindex=1}^{\scenariosnumberof} \scenarioparselement{\scenarioindex}{\elementindex}$.
In this way, the contribution of the $\elementindex$-th parameter to the overall variance (see \cref{eq:scaled variance}) only depends on $\weightpar{\elementindex}$. 
When choosing $\weightpar{1}=\ldots=\weightpar{53}$, the acceleration of the leading vehicle would contribute 50 times more to the overall variance of \cref{eq:scaled variance}, because $\numberoftimesegments=50$ elements are used to describe the acceleration. 
For the \ac{lvd} scenarios, we want to give the acceleration the same importance as each of the other parameters, so we choose $\weightpar{1}=\ldots=\weightpar{50}=1/\sqrt{\numberoftimesegments}$ and $\weightpar{51}=\weightpar{52}=\weightpar{53}=1$.

\setlength{\figurewidth}{.97\linewidth}
\setlength{\figureheight}{.7\figurewidth}

From the 289 cut-in scenarios, \SI{80}{\percent} are used for training (so $\scenariosnumberof=231$) and \SI{20}{\percent} are used for testing (so $\scenariostestnumberof=58$).
Both cut-in scenarios from the left and from the right are considered.
The training data are used for generating $\scenariosgeneratednumberof=10000$ parameter vectors that describe cut-in scenarios.
A cut-in scenario is described using the speed of the vehicle that performs the lane change and its lateral position with respect to the center of the ego vehicle's lane (so $\dimensiontimeseries=2$) at $\numberoftimesegments=50$ time instants.
In case of a cut-in scenario from the left, the lateral position is positive when the cutting-in vehicle is on the left of the center of ego vehicle's lane and vice versa for a cut-in scenario from the right.
Furthermore, $\dimensionextraparameters=3$ extra parameters are used to describe a cut-in scenario: the duration of the scenario, the initial speed of the ego vehicle, and the initial longitudinal position of the cutting-in vehicle with respect to the ego vehicle.
Thus, $\dimensionscenariopars=103$.
To give the same importance to the speed of the vehicle that performs the lane change, its lateral position, and the 3 extra parameters, the weights are calculated using \cref{eq:weights calculation} with $\weightpar{1}=\ldots=\weightpar{100}=1/\sqrt{\numberoftimesegments}$ and $\weightpar{101}=\weightpar{102}=\weightpar{103}=1$.

\subsection{Approximation of scenarios with SVD}
\label{sec:case study svd approximation}

As explained in \cref{sec:kde}, using too many parameters will lead to poor estimations of the \ac{pdf} of the parameters.
We use \iac{svd} to obtain a reduced number of parameters that best describe the original scenarios parameters.
This section illustrates the approximation of the original scenario parameters using the parameters obtained after applying the \ac{svd}.

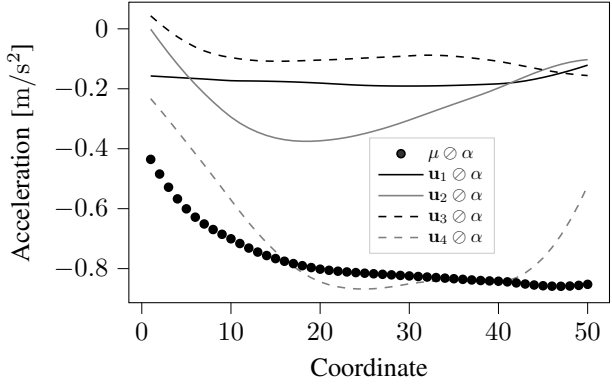
\begin{figure}
	\centering
	\input{figs/singular_vectors.tikz}
	\caption{The first $\numberoftimesegments=50$ coordinates of the first four columns of $\svdu$ after scaling with $\weights$ for the \ac{lvd} scenarios.
	Note that $\elementdivision$ denotes element-wise division.}
	\label{fig:singular vectors}
\end{figure}

Following the approximation of \cref{eq:svd approximation}, the scaled parameter vector, $\weights\elementproduct\scenariopars_{\scenarioindex}$, is approximated using a linear combination of the first $\dimension$ columns of $\svdu$, i.e., $\svduvec{1}, \ldots, \svduvec{\dimension}$.
In \cref{fig:singular vectors,tab:singular vectors}, $\svdmean$ and the first four columns of $\svdu$ are shown for the \ac{lvd} scenarios.
For an easier interpretation, the original scaling of the parameters by $\weights$ is undone via the element-wise division by $\weights$.
\Cref{fig:singular vectors} shows that the average scenario starts with a deceleration of about $\SI{0.4}{\meter\per\second\squared}$ and ends with a deceleration of about $\SI{0.8}{\meter\per\second\squared}$. 
\Cref{tab:singular vectors} shows that the average scenario duration is \SI{4.73}{\second}, the average initial speed of the leading vehicle is \SI{22.11}{\kilo\meter\per\hour}, and the average initial time gap is \SI{1.49}{\second}.
Since each scenario is estimated by combining the curves in \cref{fig:singular vectors} and values in \cref{tab:singular vectors}, it can be seen that the approximations do not contain complex acceleration curves.
In other words, the accelerations will be smoothed and the details may get lost. 
The amount of smoothing depends on $\dimension$, i.e., the number of vectors of $\svdu$ that are used to approximate the original parameter vector. 
Choosing the value of $\dimension$ is a trade-off: a higher value of $\dimension$ leads to less smoothing and, therefore, a smaller approximation error, but choosing $\dimension$ too large leads to problems when estimating the \ac{pdf} of the new parameters.

\setlength{\figurewidth}{.9\linewidth}
\setlength{\figureheight}{.7\figurewidth}

\begin{table}
	\centering
	\caption{The last $\dimensionextraparameters=3$ coordinates of the first four columns of $\svdu$ after scaling with $\weights$ for the \ac{lvd} scenarios.
	Note that $\elementdivision$ denotes element-wise division.}
	\label{tab:singular vectors}
	\begin{tabular}{cS[table-format=3.2]S[table-format=4.2]p{1.5em}S[table-format=3.2]}
		\toprule
		& {Coordinate 51} & \multicolumn{2}{c}{Coordinate 52} & {Coordinate 53} \\
		& {Scenario duration} & \multicolumn{2}{c}{Initial speed} & {Initial time gap} \\ \otoprule
		$\svdmean\elementdivision\weights$ & 4.73\,\si{\second} & 22.11\,\si{\kilo\meter\per\hour} && 1.49\,\si{\second} \\
		$\svduvec{1}\elementdivision\weights$ & -1.50\,\si{\second} & -15.17\,\si{\kilo\meter\per\hour} && 0.28\,\si{\second} \\
		$\svduvec{2}\elementdivision\weights$ & -3.09\,\si{\second} & 12.22\,\si{\kilo\meter\per\hour} && -0.06\,\si{\second} \\
		$\svduvec{3}\elementdivision\weights$ & 1.15\,\si{\second} & -16.52\,\si{\kilo\meter\per\hour} && 0.29\,\si{\second} \\
		$\svduvec{4}\elementdivision\weights$ & 1.32\,\si{\second} & -2.88\,\si{\kilo\meter\per\hour} && -0.08\,\si{\second} \\
		\bottomrule
	\end{tabular}
\end{table}

\begin{figure}
	\centering
	\input{figs/corner_cases.tikz}
	\caption{Five scenarios that require the highest average deceleration of the follower.
		The black lines denote the observed scenarios and the gray lines denote their approximations based on the $\dimension=4$ parameters.
		The corresponding initial time gaps are listed in \cref{tab:corner cases}.}
	\label{fig:corner cases}
\end{figure}
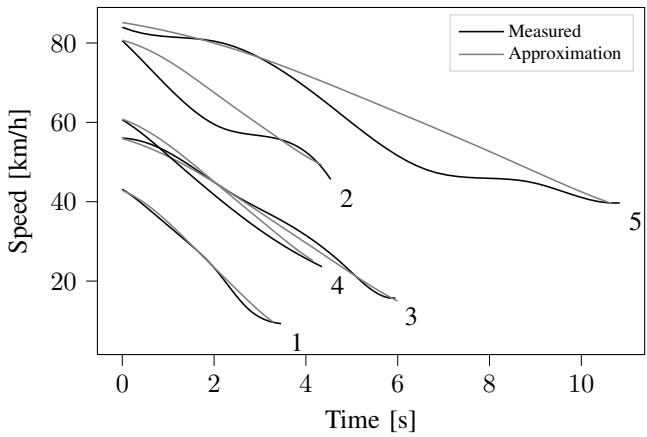

\Cref{fig:corner cases} shows five \ac{lvd} scenarios.
These selected \ac{lvd} scenarios correspond to the five \ac{lvd} scenarios that require the highest average deceleration of the following vehicle.
The line with the ``1'' denotes the \ac{lvd} scenario that requires the highest average deceleration. 
\Cref{tab:corner cases} lists the values of $\svdsv{\svdindex}\svdventry{\svdindex}{\scenarioindex}$ for $\svdindex\in\{1,\ldots,\dimension\}$ with $\dimension=4$ that are used to approximate the original scenarios according to the approximation in \cref{eq:svd approximation}.
The gray lines in \cref{fig:corner cases} show the approximated speed of the five \ac{lvd} scenarios.
\Cref{tab:corner cases} shows the initial time gaps of the five scenarios shown in \cref{fig:corner cases}.
These five scenarios illustrate that the accelerations are smoothed, but the main characteristics of the scenarios are captured by the approximations: the average deceleration, the scenario duration, the initial speed, and the initial time gap are well approximated.

\setlength{\figurewidth}{\linewidth}
\setlength{\figureheight}{.7\figurewidth}

\begin{table}
	\centering
	\caption{Initial time gaps of the five scenarios that require the highest average deceleration of the follower.
		The corresponding speeds are shown in \cref{fig:corner cases}.}
	\label{tab:corner cases}
	\begin{tabular}{crrrrcc}
		\toprule
		\# & $\svdsv{1}\svdventry{\scenarioindex}{1}$ & $\svdsv{2}\svdventry{\scenarioindex}{2}$ & $\svdsv{3}\svdventry{\scenarioindex}{3}$ & $\svdsv{4}\svdventry{\scenarioindex}{4}$ & \multicolumn{2}{c}{Initial time gap} \\
		& & & & & Original &  Approximated \\ \otoprule
		1 &   2.21 &   0.30 &  -0.03 &   2.16 & \SI{1.91}{\second} & \SI{1.91}{\second} \\
		2 &  -0.28 &   0.75 &  -0.46 &   1.56 & \SI{1.08}{\second} & \SI{1.11}{\second} \\
		3 & 0.61 &  -0.21 &  -0.42 &   1.53 & \SI{1.43}{\second} & \SI{1.43}{\second} \\
		4 & 1.74 &   0.25 &   0.58 &   1.63 & \SI{2.00}{\second} & \SI{2.00}{\second} \\
		5 & -1.74 &  -0.71 &  -0.49 &   1.30 & \SI{0.81}{\second} & \SI{0.80}{\second} \\
		\bottomrule
	\end{tabular}
\end{table}

\subsection{Generating scenario parameters}
\label{sec:case study parameter reduction}

\begin{figure}
	\centering
	\input{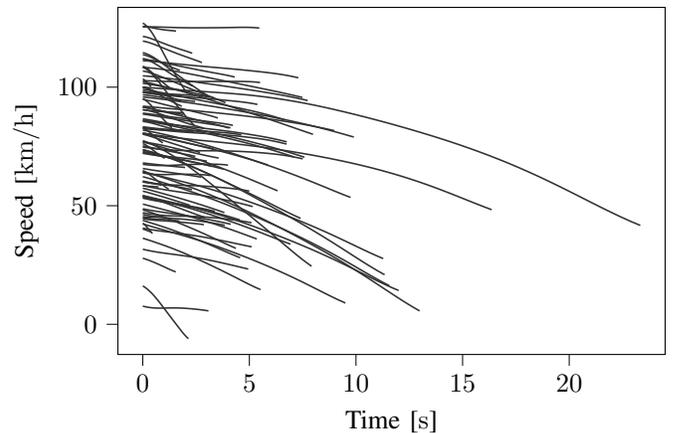}
	\caption{Speed of the leading vehicle during 100 generated \ac{lvd} scenarios.}
	\label{fig:speed lvd generated}
\end{figure}

An important parameter for the generation of the scenario parameter vectors is the number of reduced parameters ($\dimension$).
One approach is to look at the so-called explained variance of \cref{eq:explained variance} of the first $\dimension$ singular values, see \cref{tab:explained variance}.
The first 4 singular values already explain \SI{90.4}{\percent} of the variance for the \ac{lvd} scenarios, so $\dimension=4$ might be a suitable choice.
In \cref{fig:speed lvd generated}, the speed of the leading vehicle of 100 generated \ac{lvd} scenarios is shown using $\dimension=4$.

\begin{table}
	\centering
	\caption{Explained variance according to \cref{eq:explained variance}.}
	\label{tab:explained variance}
	\begin{tabular}{lcc}
		\toprule
		$\dimension$  &      \Acl{lvd}      &       Cut-in        \\
		\otoprule
		1             & \SI{36.9}{\percent} & \SI{36.9}{\percent} \\
		2             & \SI{63.0}{\percent} & \SI{63.3}{\percent} \\
		3             & \SI{78.0}{\percent} & \SI{84.5}{\percent} \\
		4             & \SI{90.4}{\percent} & \SI{94.1}{\percent} \\
		5             & \SI{94.5}{\percent} & \SI{96.9}{\percent} \\
		6             & \SI{96.7}{\percent} & \SI{99.0}{\percent} \\
		7             & \SI{98.2}{\percent} & \SI{99.6}{\percent} \\
		8             & \SI{99.2}{\percent} & \SI{99.8}{\percent} \\ \bottomrule
	\end{tabular}
\end{table}

\begin{figure}
	\centering
	\input{figs/scores_1.tikz}
	\caption{Medians of the metrics for the set of generated \ac{lvd} scenario parameters. 
	Note that $\dimension=1$ is excluded, because its metrics are an order of magnitude higher than for $\dimension=2$ and would, therefore, not be visible with the current scaling of the y-axis.}
	\label{fig:scores lvd}
\end{figure}
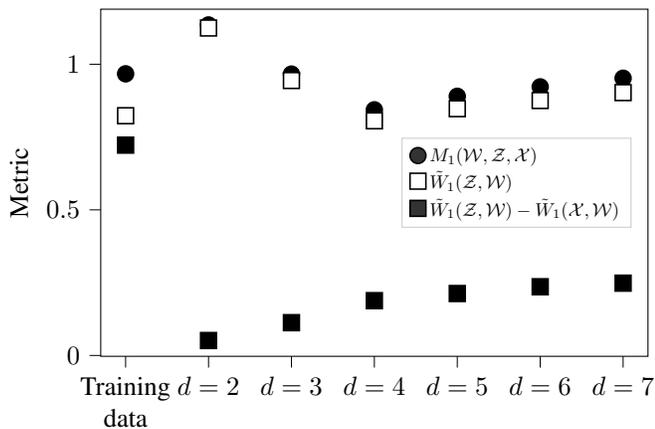

Another way to determine $\dimension$ is to use the \ac{sr} metric $\proposedmetric{\wassersteincoefficient}{\scenariosetgenerated}{\scenariosettest}{\scenarioset}$ defined in \cref{eq:proposed metric}.
In \cref{fig:scores lvd}, the result is shown when applying this metric with $\wassersteincoefficient=1$, alongside with the empirical Wasserstein metric $\wassersteinemp{1}{\scenariosettest}{\scenariosetgenerated}$  and the penalty $\wassersteinemp{1}{\scenariosettest}{\scenariosetgenerated} - \wassersteinemp{1}{\scenarioset}{\scenariosetgenerated}$. 
Each point in \cref{fig:scores lvd} represents the median\footnote{We preferred to use the median instead of the mean, such that the result is less influenced by outliers \autocite{doerr2019resampling}.} when applying the metric 200 times, each time with a different (random) partition of the training data $\scenarioset$ and test data $\scenariosettest$.
The standard deviation of the medians in \cref{fig:scores lvd}, estimated using bootstrapping \autocite{efron1992bootstrap}, is 0.005 or less.
For the \ac{sr} metric, the penalty is weighted using $\penaltyweight=0.25$.
The choice of $\penaltyweight=0.25$ is justified in \cref{sec:case study comparison metric}.

\setlength{\figureheight}{.7\figurewidth}

The most left points in \cref{fig:scores lvd} represent the metric in case the set of training data $\scenarioset$ is directly used to sample the scenario parameters instead of the approach of \cref{sec:generation}. 
Here, $\scenariosetgenerated$ is a selection with replacement of $\scenariosgeneratednumberof$ scenarios from $\scenarioset$, i.e.:
\begin{equation}
	\scenarioparsgenerated_{\scenarioindex} = \scenariopars_{\floor{\uniformvariable}},
	\uniformvariable \sim \uniformdist{1}{\scenariosnumberof+1},
	\forall \scenarioindex \in \{1,\ldots,\scenariosgeneratednumberof\},
\end{equation}
where $\uniformdist{1}{\scenariosnumberof+1}$ denotes the continuous uniform distribution with boundaries $1$ and $\scenariosnumberof+1$, and $\floor{\cdot}$ denotes the floor function.
Using the training data directly for ``generating scenarios'' leads to a low empirical Wasserstein metric.
The downside is that there is not much variation among the generated scenarios.
Therefore, the penalty is also the highest, which results in $\proposedmetric{1}{\scenariosetgenerated}{\scenariosettest}{\scenarioset}\approx0.967$.
Looking at $\dimension=4$, the empirical Wasserstein metric (open squares) is approximately similar compared to when the training set is directly used.
Due to the sampling of the scenario parameters from the \ac{kde}, the generated scenarios contain more variation than the training set, resulting in a lower penalty and, therefore, a lower metric evaluation of $\proposedmetric{1}{\scenariosetgenerated}{\scenariosettest}{\scenarioset}\approx0.843$.
Increasing $\dimension$ even further results in higher metric evaluations.
So based on the proposed metric, $\dimension=4$ seems the right choice.

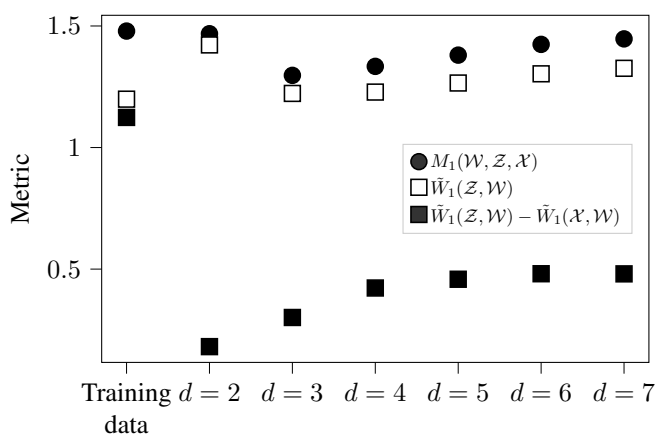
\begin{figure}
	\centering
	\input{figs/scores_2.tikz}
	\caption{Medians of the metrics for the set of generated cut-in scenario parameters. 
		}
	\label{fig:scores ci}
\end{figure}

\cref{fig:scores ci} shows the results of the generation of the cut-in scenario parameters in a similar way as \cref{fig:scores lvd}.
The standard deviation of all points in \cref{fig:scores ci} is less than 0.008.
The lowest penalty is obtained with $\dimension=2$, but the higher empirical Wasserstein distance suggests that too much information is lost.
The best result, i.e., where the \ac{sr} metric, $\proposedmetric{1}{\scenariosetgenerated}{\scenariosettest}{\scenarioset}$, is minimal, is obtained at $\dimension=3$. 

\subsection{Comparison with other approaches}
\label{sec:case study comparison method}

Our proposed method utilizes \iac{svd} to obtain the scenario parameters and multivariate \ac{kde} to estimate the \ac{pdf} of these parameters.
To illustrate the advantages of these choices, the results of our method are compared with alternative approaches. 
First, instead of using \iac{svd} for obtaining the parameters, a fixed parameterization is used, such as in \autocite{deGelder2017assessment, zofka2015datadrivetrafficscenarios, thal2020incorporating}.
Second, instead of using \ac{kde} to estimate the \ac{pdf} of the parameters, a Gaussian distribution like in \autocite{gietelink2007phd} is assumed.
Third, the parameters are assumed to be independent. 

When using a fixed parameterization for the \ac{lvd} scenario, 4 parameters describe the scenario \autocite{deGelder2017assessment}: the speed reduction of the leading vehicle, the final speed of the leading vehicle, the duration of the scenario, and the initial time gap between the leading vehicle and the ego vehicle.
The speed of the leading vehicle is assumed to follow a sinusoidal function, such that the acceleration at the start and at the end of the scenario equals zero.
In case of the cut-in scenario, 5 parameters describe the scenario: the mean speed of the vehicle cutting in, its initial lateral position with respect to the center of the ego vehicle's lane, the duration of the scenario, the initial speed of the ego vehicle, and the initial longitudinal position of the vehicle cutting in with respect to the ego vehicle.
The speed of the vehicle cutting in is assumed to be constant. 
Its lateral position is assumed to follow a sinusoidal function, such that the vehicle ends at the center of the ego vehicle's lane.
For estimating the \ac{pdf} of these parameters, the comparison considers 4 possibilities: multivariate \ac{kde}, multiple univariate \acp{kde}, a multivariate Gaussian distribution, and multiple univariate Gaussian distributions.

\Cref{tab:comparison methods} shows the results of the different approaches for generating scenario parameters.
For both the \ac{lvd} scenarios, our proposed approach (top row in \cref{tab:comparison methods}) resulted in the lowest $\proposedmetric{1}{\scenariosetgenerated}{\scenariosettest}{\scenarioset}$.
For the cut-in scenarios, it is interesting to note that the scores are not very different if \ac{svd} is used to obtain the parameters.
This is partly explained by the smaller data set, because this results in a higher bandwidth\footnote{On average, the bandwidth is about 1.5 to 2 times larger for the cut-in scenarios compared to the \ac{lvd} scenarios.} that makes the \ac{kde} result with the Gaussian kernel look more like a Gaussian distribution.
Using \ac{svd} and \ac{kde} while assuming that the parameters are independent, results in an even better result: 1.28 instead of 1.30 (with a standard deviation of 0.005).
This indicates that assuming that the 3 parameters obtained with the \ac{svd} are independent, is acceptable.

\begin{table}
	\centering
	\caption{Medians of the metric $\proposedmetric{1}{\scenariosetgenerated}{\scenariosettest}{\scenarioset}$ with different approaches for generating scenarios.}
	\label{tab:comparison methods}
	\begin{tabular}{cccrr}
		\toprule
		Parameters & Distribution & Dependency  & \acs{lvd}            & Cut-in \\ \otoprule
		\acs{svd}  &  \acs{kde}   &  Dependent  & 0.84 &   1.30 \\
		\acs{svd}  &   Gaussian   &  Dependent  & 1.00 &   1.33 \\
		\acs{svd}  &  \acs{kde}   & Independent & 0.99 &   1.28 \\
		\acs{svd}  &   Gaussian   & Independent & 1.00 &   1.33 \\
		Fixed      &  \acs{kde}   &  Dependent  & 2.65 &   1.70 \\
		Fixed      &   Gaussian   &  Dependent  & 2.58 &   1.71 \\
		Fixed      &  \acs{kde}   & Independent & 2.31 &   1.67 \\
		Fixed      &   Gaussian   & Independent & 4.76 &   1.69 \\ 
		\bottomrule
	\end{tabular}
\end{table}

\subsection{Evaluating the scenario representativeness metric}
\label{sec:case study comparison metric}

To determine whether our proposed metric \cref{eq:proposed metric} correlates better with the Wasserstein metric \cref{eq:wasserstein} than the empirical Wasserstein metric \cref{eq:empirical density}, the Wasserstein metric \cref{eq:wasserstein} needs to be known. 
This is not possible because the true underlying distribution of the data is unknown.
To estimate the Wasserstein metric \cref{eq:wasserstein}, the empirical Wasserstein metric \cref{eq:empirical density} can be used with large numbers of test scenarios and generated scenario parameters, i.e., with large values of $\scenariostestnumberof$ and $\scenariosgeneratednumberof$, respectively.
Since a large number of test scenarios is not available to us, we assume a certain distribution for $\density{\cdot}$ from which the training data and the test data are generated.
The approach is as follows (the numbers are for the \ac{lvd} scenarios and, in parenthesis, the numbers for the cut-in scenarios are shown):
\begin{enumerate}
	\item Based on the original 1150 (289) scenarios, obtained from the data, the following sets of scenario parameters are generated using the proposed approach explained in \cref{sec:generation} with $\dimension=4$ ($\dimension=3$):
	\begin{itemize}
		\item A new set of training data $\scenariosetfake$ of size $\scenariosnumberof=920$ ($\scenariosnumberof=231$);
		\item A new set of test data $\scenariosettestfakesmall$ of size $\scenariostestnumberof=230$ ($\scenariostestnumberof=58$); and
		\item A large set of test data $\scenariosettestfakelarge$ of size $\scenariostestnumberof=10000$ ($\scenariostestnumberof=10000$).
	\end{itemize}
	\item Based on $\scenariosetfake$, $\scenariosgeneratednumberof=10000$ ($\scenariosgeneratednumberof=10000$) scenario parameters are generated and collected in a set $\scenariosetgeneratedfake$.
	\item Our proposed metric is computed using $\scenariosetgeneratedfake$, $\scenariosettestfakesmall$, and $\scenariosetfake$: $\proposedmetric{1}{\scenariosetgeneratedfake}{\scenariosettestfakesmall}{\scenariosetfake}$ with $\penaltyweight=0.25$.
	\item Estimate the Wasserstein metric of \cref{eq:wasserstein} using the empirical Wasserstein metric of \cref{eq:empirical wasserstein} with $\scenariosetgeneratedfake$ and $\scenariosettestfakelarge$.
	Note: to approximate the Wasserstein metric of \cref{eq:wasserstein} using the empirical Wasserstein metric of \cref{eq:empirical wasserstein}, both $\scenariosetgeneratedfake$ and $\scenariosettestfakelarge$ need to be large (but not necessarily the same) in size.
\end{enumerate}

\begin{figure}
	\centering
	\input{figs/wasserstein_2.tikz}
	\caption{Medians of the metrics for the set of $\scenariosgeneratednumberof=10000$ generated cut-in scenario parameter vectors. 
	}
	\label{fig:wasserstein comparison 2}
\end{figure}
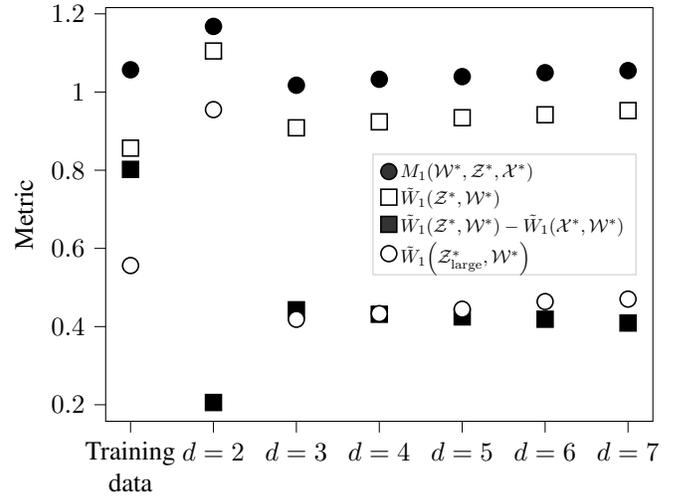

We have repeated this approach 200 times, each time with a different (random) partition of the training data $\scenarioset$ and test data $\scenariosettest$.
\cref{fig:wasserstein comparison 1,fig:wasserstein comparison 2} show the result of this approach for the \ac{lvd} scenarios and cut-in scenarios, respectively.
In both cases, the empirical Wasserstein metric $\wassersteinemp{1}{\scenariosettestfakesmall}{\scenariosetgeneratedfake}$ is minimal when the training data are directly used for the generated scenario parameters.
Thus, the empirical Wasserstein metric suggests that the best approach for generating new scenario parameters is to simply sample parameters from the training data.
The actual Wasserstein metric, estimated using $\wassersteinemp{1}{\scenariosettestfakelarge}{\scenariosetgeneratedfake}$ shows that using our proposed method outperforms sampling parameters directly from the training data.

\setlength{\figureheight}{.8\figurewidth}
\begin{figure}
	\centering
	\input{figs/wasserstein_1.tikz}
	\caption{Medians of the metrics for the set of $\scenariosgeneratednumberof=10000$ generated \ac{lvd} scenario parameter vectors. 
	In this case, the $\scenariosnumberof=920$ scenarios of $\scenariosetfake$ are sampled from $\densityestkde{\bandwidthmatrix}{\cdot}$ of \cref{eq:density est kde}, where $\densityestkde{\bandwidthmatrix}{\cdot}$ is based on the original data set $\scenarioset$.}
	\label{fig:wasserstein comparison 1}
\end{figure}
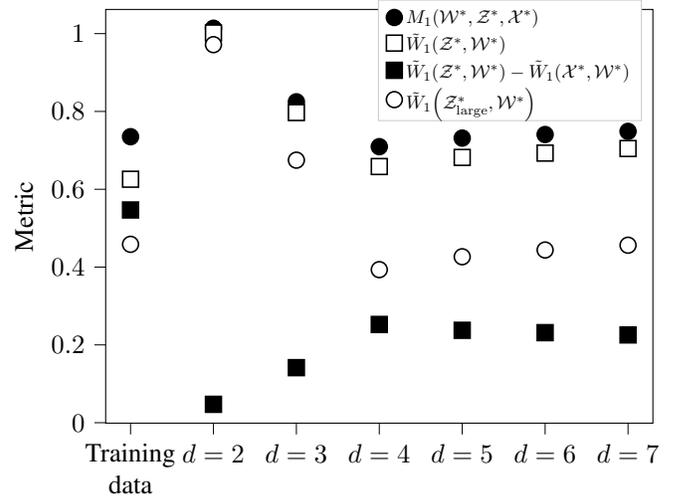

\begin{figure}
	\centering
	\input{figs/correlation.tikz}
	\caption{Correlation between the medians of $\proposedmetric{1}{\scenariosetgeneratedfake}{\scenariosettestfakesmall}{\scenariosetfake}$ (open circles in \cref{fig:wasserstein comparison 1,fig:wasserstein comparison 2}) and $\wassersteinemp{1}{\scenariosettestfakelarge}{\scenariosetgeneratedfake}$ (filled circles in \cref{fig:wasserstein comparison 1,fig:wasserstein comparison 2}) for different values of $\penaltyweight$. 
		The solid line shows the result for the \ac{lvd} scenarios with a maximum correlation of 0.992 at $\penaltyweight\approx 0.21$.
		The dashed line shows the result for the cut-in scenarios with a maximum correlation of 0.987 at $\penaltyweight\approx 0.25$.}
	\label{fig:correlation}
\end{figure}
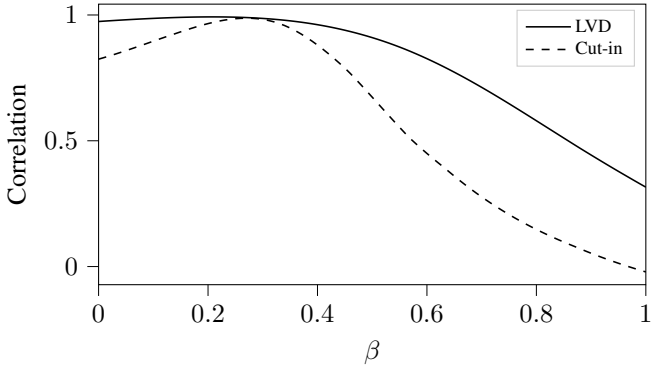

To justify the choice of $\penaltyweight=0.25$, \cref{fig:correlation} shows the correlation between the medians of the proposed metric $\proposedmetric{1}{\scenariosetgeneratedfake}{\scenariosettestfakesmall}{\scenariosetfake}$ and $\wassersteinemp{1}{\scenariosettestfakelarge}{\scenariosetgeneratedfake}$ for different values of $\penaltyweight$.
With $\penaltyweight=0$, i.e., $\proposedmetric{1}{\scenariosetgeneratedfake}{\scenariosettestfakesmall}{\scenariosetfake}=\wassersteinemp{1}{\scenariosettestfakesmall}{\scenariosetgeneratedfake}$, the correlation is 0.974 for the \ac{lvd} scenarios and 0.824 for the cut-in scenarios.
The correlation increases with increasing $\penaltyweight$ until the maximum is obtained at $\penaltyweight\approx 0.21$ for the \ac{lvd} scenarios and at $\penaltyweight\approx0.27$ for the cut-in scenarios.
The correlations at these maxima are 0.992 and 0.987, respectively.
Increasing $\penaltyweight$ further results in a lower correlation, which suggests that a choice of $\beta=0.25$ seems appropriate. 

\setlength{\figureheight}{.6\figurewidth}

The experiment described in this section can be used to determine both $\dimension$ and $\penaltyweight$ in an iterative manner given an initial choice for $\penaltyweight$ (denoted by $\penaltyweight_0$):
\begin{enumerate}
	\item Set $i=0$.
	\item Determine $\dimension_i$, i.e., the optimal number of parameters that minimizes $\proposedmetric{1}{\scenariosetgenerated}{\scenariosettest}{\scenarioset}$ using $\penaltyweight=\penaltyweight_i$.
	\item Generate $\scenariosetfake$, $\scenariosetgeneratedfake$, $\scenariosettestfakesmall$, and $\scenariosettestfakelarge$ using the approach described in this section with $\dimension=\dimension_i$.
	\item Increase $i$ by 1.
	\item Determine $\penaltyweight_i$ by maximizing the correlation between $\proposedmetric{1}{\scenariosetgeneratedfake}{\scenariosettestfakesmall}{\scenariosetfake}$ with $\penaltyweight=\penaltyweight_i$ and $\wassersteinemp{1}{\scenariosettestfakelarge}{\scenariosetgeneratedfake}$ (e.g., see \cref{fig:correlation}).
	\item Repeat step 2.
	\item Stop if $\dimension_i = \dimension_{i-1}$. Otherwise, return to step 3.
\end{enumerate}
As an initial choice, $\penaltyweight_0=0.25$ seems appropriate.
More specifically, when choosing $\penaltyweight_0\in[0.1, 1]$, the optimal choice of $\dimension$ is found after one iteration.

%% file: figs/lead_braking.tikz
\begin{tikzpicture}
\definecolor{color0}{rgb}{0.8,1,0.8}
\definecolor{color1}{rgb}{0,0.4375,0.75}

\begin{axis}[
xmin=-15, xmax=15,
ymin=-5, ymax=5,
width=\figurewidth,
height=0.33\figurewidth,
tick align=outside,
xticklabel style = {align=center,text width=1},
yticklabel style = {align=right,text width=1},
tick pos=left,
x grid style={white!69.01960784313725!black},
y grid style={white!69.01960784313725!black},
axis background/.style={fill=color0},
ticks=none,
scale only axis
]
\path [draw=white!80.0!black, fill=white!80.0!black] (axis cs:-15,2.8)
--(axis cs:15,2.8)
--(axis cs:15,-2.8)
--(axis cs:-15,-2.8)
--cycle;

\addplot [semithick, black, forget plot]
table {%
-15 2.8
15 2.8
};
\addplot [semithick, black, forget plot]
table {%
15 -2.8
-15 -2.8
};
\addplot [semithick, white, forget plot]
table {%
-15 0
-14.5 0
};
\addplot [semithick, white, forget plot]
table {%
-12.5 0
-11.5 0
};
\addplot [semithick, white, forget plot]
table {%
-9.5 0
-8.5 0
};
\addplot [semithick, white, forget plot]
table {%
-6.5 0
-5.5 0
};
\addplot [semithick, white, forget plot]
table {%
-3.5 0
-2.5 0
};
\addplot [semithick, white, forget plot]
table {%
-0.499999999999999 0
0.500000000000002 0
};
\addplot [semithick, white, forget plot]
table {%
2.5 0
3.5 0
};
\addplot [semithick, white, forget plot]
table {%
5.5 0
6.5 0
};
\addplot [semithick, white, forget plot]
table {%
8.5 0
9.5 0
};
\addplot [semithick, white, forget plot]
table {%
11.5 0
12.5 0
};
\addplot [semithick, white, forget plot]
table {%
14.5 0
15 0
};
\path [fill=black, draw opacity=0] (axis cs:5,0.86)
--(axis cs:0.75,0.86)
--(axis cs:0.75,0.59)
--(axis cs:5,0.59)
--cycle;

\path [fill=black, draw opacity=0] (axis cs:5,1.94)
--(axis cs:0.75,1.94)
--(axis cs:0.75,2.21)
--(axis cs:5,2.21)
--cycle;

\path [draw=black, fill=color1] (axis cs:-12.25,1.40401785714286)
--(axis cs:-12.2418918918919,1.92633928571429)
--(axis cs:-12.1851351351351,2.12723214285714)
--(axis cs:-12.1040540540541,2.19955357142857)
--(axis cs:-11.9743243243243,2.24776785714286)
--(axis cs:-11.5121621621622,2.30401785714286)
--(axis cs:-8.09864864864865,2.26383928571429)
--(axis cs:-8.00945945945946,2.20758928571429)
--(axis cs:-7.87162162162162,2.03080357142857)
--(axis cs:-7.81486486486487,1.83794642857143)
--(axis cs:-7.75,1.40401785714286)
--(axis cs:-7.75,1.39598214285714)
--(axis cs:-7.81486486486487,0.962053571428571)
--(axis cs:-7.87162162162162,0.769196428571429)
--(axis cs:-8.00945945945946,0.592410714285714)
--(axis cs:-8.09864864864865,0.536160714285714)
--(axis cs:-11.5121621621622,0.495982142857143)
--(axis cs:-11.9743243243243,0.552232142857143)
--(axis cs:-12.1040540540541,0.600446428571428)
--(axis cs:-12.1851351351351,0.672767857142857)
--(axis cs:-12.2418918918919,0.873660714285714)
--(axis cs:-12.25,1.39598214285714)
--cycle;

\path [draw=black, fill=color1] (axis cs:-9.33108108108108,2.23169642857143)
--(axis cs:-9.33918918918919,2.43258928571429)
--(axis cs:-9.33108108108108,2.48080357142857)
--(axis cs:-9.29864864864865,2.45669642857143)
--(axis cs:-9.25,2.24776785714286)
--cycle;

\path [draw=black, fill=color1] (axis cs:-9.33108108108108,0.568303571428571)
--(axis cs:-9.33918918918919,0.367410714285714)
--(axis cs:-9.33108108108108,0.319196428571428)
--(axis cs:-9.29864864864865,0.343303571428571)
--(axis cs:-9.25,0.552232142857143)
--cycle;

\path [draw=black, fill=white] (axis cs:-11.7148648648649,1.40401785714286)
--(axis cs:-11.6986486486486,1.77366071428571)
--(axis cs:-11.65,1.99866071428571)
--(axis cs:-11.5851351351351,2.11919642857143)
--(axis cs:-11.5283783783784,2.12723214285714)
--(axis cs:-11.0175675675676,1.990625)
--(axis cs:-11.0662162162162,1.85401785714286)
--(axis cs:-11.0743243243243,1.64508928571429)
--(axis cs:-11.0743243243243,1.15491071428571)
--(axis cs:-11.0662162162162,0.945982142857143)
--(axis cs:-11.0175675675676,0.809375)
--(axis cs:-11.5283783783784,0.672767857142857)
--(axis cs:-11.5851351351351,0.680803571428571)
--(axis cs:-11.65,0.801339285714285)
--(axis cs:-11.6986486486486,1.02633928571429)
--(axis cs:-11.7148648648649,1.39598214285714)
--cycle;

\path [draw=black, fill=white] (axis cs:-8.91756756756757,1.40401785714286)
--(axis cs:-8.94189189189189,1.83794642857143)
--(axis cs:-9.03108108108108,2.11116071428571)
--(axis cs:-9.08783783783784,2.16741071428571)
--(axis cs:-9.61486486486486,1.95848214285714)
--(axis cs:-9.56621621621622,1.765625)
--(axis cs:-9.55,1.596875)
--(axis cs:-9.55,1.203125)
--(axis cs:-9.56621621621622,1.034375)
--(axis cs:-9.61486486486486,0.841517857142857)
--(axis cs:-9.08783783783784,0.632589285714286)
--(axis cs:-9.03108108108108,0.688839285714286)
--(axis cs:-8.94189189189189,0.962053571428571)
--(axis cs:-8.91756756756757,1.39598214285714)
--cycle;

\path [draw=black, fill=white] (axis cs:-11.0256756756757,2.23169642857143)
--(axis cs:-10.8148648648649,2.23169642857143)
--(axis cs:-10.8148648648649,2.07901785714286)
--(axis cs:-10.9202702702703,2.12723214285714)
--cycle;

\path [draw=black, fill=white] (axis cs:-11.0256756756757,0.568303571428571)
--(axis cs:-10.8148648648649,0.568303571428571)
--(axis cs:-10.8148648648649,0.720982142857143)
--(axis cs:-10.9202702702703,0.672767857142857)
--cycle;

\path [draw=black, fill=white] (axis cs:-10.7662162162162,2.06294642857143)
--(axis cs:-10.7662162162162,2.19151785714286)
--(axis cs:-10.7337837837838,2.22366071428571)
--(axis cs:-10.2067567567568,2.22366071428571)
--(axis cs:-10.1824324324324,2.18348214285714)
--(axis cs:-10.2310810810811,2.03883928571429)
--(axis cs:-10.3040540540541,2.00669642857143)
--(axis cs:-10.5716216216216,2.02276785714286)
--cycle;

\path [draw=black, fill=white] (axis cs:-10.7662162162162,0.737053571428571)
--(axis cs:-10.7662162162162,0.608482142857143)
--(axis cs:-10.7337837837838,0.576339285714286)
--(axis cs:-10.2067567567568,0.576339285714286)
--(axis cs:-10.1824324324324,0.616517857142857)
--(axis cs:-10.2310810810811,0.761160714285714)
--(axis cs:-10.3040540540541,0.793303571428571)
--(axis cs:-10.5716216216216,0.777232142857143)
--cycle;

\path [draw=black, fill=white] (axis cs:-10.15,1.98258928571429)
--(axis cs:-10.0202702702703,2.23169642857143)
--(axis cs:-9.28243243243243,2.23169642857143)
--(axis cs:-9.29054054054054,2.18348214285714)
--(axis cs:-9.70405405405405,2.00669642857143)
--cycle;

\path [draw=black, fill=white] (axis cs:-10.15,0.817410714285714)
--(axis cs:-10.0202702702703,0.568303571428571)
--(axis cs:-9.28243243243243,0.568303571428571)
--(axis cs:-9.29054054054054,0.616517857142857)
--(axis cs:-9.70405405405405,0.793303571428571)
--cycle;

\path [draw=black, fill=white] (axis cs:-8.2527027027027,2.24776785714286)
--(axis cs:-8.09054054054054,2.23973214285714)
--(axis cs:-7.96891891891892,2.11919642857143)
--(axis cs:-7.91216216216216,2.02276785714286)
--(axis cs:-7.89594594594595,1.92633928571429)
--(axis cs:-7.89594594594595,1.75758928571429)
--(axis cs:-7.98513513513513,1.91026785714286)
--cycle;

\path [draw=black, fill=white] (axis cs:-8.2527027027027,0.552232142857143)
--(axis cs:-8.09054054054054,0.560267857142857)
--(axis cs:-7.96891891891892,0.680803571428571)
--(axis cs:-7.91216216216216,0.777232142857143)
--(axis cs:-7.89594594594595,0.873660714285714)
--(axis cs:-7.89594594594595,1.04241071428571)
--(axis cs:-7.98513513513513,0.889732142857143)
--cycle;

\path [draw=black, fill=red] (axis cs:2.75,1.40401785714286)
--(axis cs:2.75810810810811,1.92633928571429)
--(axis cs:2.81486486486487,2.12723214285714)
--(axis cs:2.89594594594595,2.19955357142857)
--(axis cs:3.02567567567568,2.24776785714286)
--(axis cs:3.48783783783784,2.30401785714286)
--(axis cs:6.90135135135135,2.26383928571429)
--(axis cs:6.99054054054054,2.20758928571429)
--(axis cs:7.12837837837838,2.03080357142857)
--(axis cs:7.18513513513513,1.83794642857143)
--(axis cs:7.25,1.40401785714286)
--(axis cs:7.25,1.39598214285714)
--(axis cs:7.18513513513513,0.962053571428572)
--(axis cs:7.12837837837838,0.769196428571429)
--(axis cs:6.99054054054054,0.592410714285714)
--(axis cs:6.90135135135135,0.536160714285714)
--(axis cs:3.48783783783784,0.495982142857143)
--(axis cs:3.02567567567568,0.552232142857143)
--(axis cs:2.89594594594595,0.600446428571429)
--(axis cs:2.81486486486487,0.672767857142857)
--(axis cs:2.75810810810811,0.873660714285714)
--(axis cs:2.75,1.39598214285714)
--cycle;

\path [draw=black, fill=red] (axis cs:5.66891891891892,2.23169642857143)
--(axis cs:5.66081081081081,2.43258928571429)
--(axis cs:5.66891891891892,2.48080357142857)
--(axis cs:5.70135135135135,2.45669642857143)
--(axis cs:5.75,2.24776785714286)
--cycle;

\path [draw=black, fill=red] (axis cs:5.66891891891892,0.568303571428571)
--(axis cs:5.66081081081081,0.367410714285714)
--(axis cs:5.66891891891892,0.319196428571429)
--(axis cs:5.70135135135135,0.343303571428571)
--(axis cs:5.75,0.552232142857143)
--cycle;

\path [draw=black, fill=white] (axis cs:3.28513513513514,1.40401785714286)
--(axis cs:3.30135135135135,1.77366071428571)
--(axis cs:3.35,1.99866071428571)
--(axis cs:3.41486486486486,2.11919642857143)
--(axis cs:3.47162162162162,2.12723214285714)
--(axis cs:3.98243243243243,1.990625)
--(axis cs:3.93378378378378,1.85401785714286)
--(axis cs:3.92567567567568,1.64508928571429)
--(axis cs:3.92567567567568,1.15491071428571)
--(axis cs:3.93378378378378,0.945982142857143)
--(axis cs:3.98243243243243,0.809375)
--(axis cs:3.47162162162162,0.672767857142857)
--(axis cs:3.41486486486486,0.680803571428571)
--(axis cs:3.35,0.801339285714286)
--(axis cs:3.30135135135135,1.02633928571429)
--(axis cs:3.28513513513514,1.39598214285714)
--cycle;

\path [draw=black, fill=white] (axis cs:6.08243243243243,1.40401785714286)
--(axis cs:6.05810810810811,1.83794642857143)
--(axis cs:5.96891891891892,2.11116071428571)
--(axis cs:5.91216216216216,2.16741071428571)
--(axis cs:5.38513513513514,1.95848214285714)
--(axis cs:5.43378378378378,1.765625)
--(axis cs:5.45,1.596875)
--(axis cs:5.45,1.203125)
--(axis cs:5.43378378378378,1.034375)
--(axis cs:5.38513513513514,0.841517857142857)
--(axis cs:5.91216216216216,0.632589285714286)
--(axis cs:5.96891891891892,0.688839285714286)
--(axis cs:6.05810810810811,0.962053571428572)
--(axis cs:6.08243243243243,1.39598214285714)
--cycle;

\path [draw=black, fill=white] (axis cs:3.97432432432432,2.23169642857143)
--(axis cs:4.18513513513513,2.23169642857143)
--(axis cs:4.18513513513513,2.07901785714286)
--(axis cs:4.07972972972973,2.12723214285714)
--cycle;

\path [draw=black, fill=white] (axis cs:3.97432432432432,0.568303571428571)
--(axis cs:4.18513513513513,0.568303571428571)
--(axis cs:4.18513513513513,0.720982142857143)
--(axis cs:4.07972972972973,0.672767857142857)
--cycle;

\path [draw=black, fill=white] (axis cs:4.23378378378378,2.06294642857143)
--(axis cs:4.23378378378378,2.19151785714286)
--(axis cs:4.26621621621622,2.22366071428571)
--(axis cs:4.79324324324324,2.22366071428571)
--(axis cs:4.81756756756757,2.18348214285714)
--(axis cs:4.76891891891892,2.03883928571429)
--(axis cs:4.69594594594595,2.00669642857143)
--(axis cs:4.42837837837838,2.02276785714286)
--cycle;

\path [draw=black, fill=white] (axis cs:4.23378378378378,0.737053571428572)
--(axis cs:4.23378378378378,0.608482142857143)
--(axis cs:4.26621621621622,0.576339285714286)
--(axis cs:4.79324324324324,0.576339285714286)
--(axis cs:4.81756756756757,0.616517857142857)
--(axis cs:4.76891891891892,0.761160714285714)
--(axis cs:4.69594594594595,0.793303571428572)
--(axis cs:4.42837837837838,0.777232142857143)
--cycle;

\path [draw=black, fill=white] (axis cs:4.85,1.98258928571429)
--(axis cs:4.97972972972973,2.23169642857143)
--(axis cs:5.71756756756757,2.23169642857143)
--(axis cs:5.70945945945946,2.18348214285714)
--(axis cs:5.29594594594595,2.00669642857143)
--cycle;

\path [draw=black, fill=white] (axis cs:4.85,0.817410714285714)
--(axis cs:4.97972972972973,0.568303571428571)
--(axis cs:5.71756756756757,0.568303571428571)
--(axis cs:5.70945945945946,0.616517857142857)
--(axis cs:5.29594594594595,0.793303571428572)
--cycle;

\path [draw=black, fill=white] (axis cs:6.7472972972973,2.24776785714286)
--(axis cs:6.90945945945946,2.23973214285714)
--(axis cs:7.03108108108108,2.11919642857143)
--(axis cs:7.08783783783784,2.02276785714286)
--(axis cs:7.10405405405405,1.92633928571429)
--(axis cs:7.10405405405405,1.75758928571429)
--(axis cs:7.01486486486487,1.91026785714286)
--cycle;

\path [draw=black, fill=white] (axis cs:6.7472972972973,0.552232142857143)
--(axis cs:6.90945945945946,0.560267857142857)
--(axis cs:7.03108108108108,0.680803571428572)
--(axis cs:7.08783783783784,0.777232142857143)
--(axis cs:7.10405405405405,0.873660714285714)
--(axis cs:7.10405405405405,1.04241071428571)
--(axis cs:7.01486486486487,0.889732142857143)
--cycle;

\addplot [black, forget plot]
table {%
-11.6337837837838 2.14330357142857
-11.8608108108108 2.13526785714286
-12.0716216216216 2.07901785714286
-12.1364864864865 1.99866071428571
-12.1364864864865 0.801339285714285
-12.0716216216216 0.720982142857143
-11.8608108108108 0.664732142857143
-11.6337837837838 0.656696428571428
};
\addplot [black, forget plot]
table {%
-8.00945945945946 1.82991071428571
-7.98513513513513 1.85401785714286
-7.89594594594595 1.71741071428571
-7.89594594594595 1.08258928571429
-7.98513513513513 0.945982142857143
-8.00945945945946 0.970089285714286
};
\addplot [black, forget plot]
table {%
-8.26081081081081 2.13526785714286
-9.0472972972973 2.16741071428571
-8.00945945945946 1.82991071428571
-8.00945945945946 0.970089285714286
-9.0472972972973 0.632589285714286
-8.26081081081081 0.664732142857143
};
\addplot [semithick, red, forget plot]
table {%
-7.75 1.4
-2.75 1.4
};
\addplot [semithick, red, forget plot]
table {%
-3.5 2.15
-2.75 1.4
-3.5 0.65
};
\addplot [black, forget plot]
table {%
3.36621621621622 2.14330357142857
3.13918918918919 2.13526785714286
2.92837837837838 2.07901785714286
2.86351351351351 1.99866071428571
2.86351351351351 0.801339285714286
2.92837837837838 0.720982142857143
3.13918918918919 0.664732142857143
3.36621621621622 0.656696428571429
};
\addplot [black, forget plot]
table {%
6.99054054054054 1.82991071428571
7.01486486486487 1.85401785714286
7.10405405405405 1.71741071428571
7.10405405405405 1.08258928571429
7.01486486486487 0.945982142857143
6.99054054054054 0.970089285714286
};
\addplot [black, forget plot]
table {%
6.73918918918919 2.13526785714286
5.9527027027027 2.16741071428571
6.99054054054054 1.82991071428571
6.99054054054054 0.970089285714286
5.9527027027027 0.632589285714286
6.73918918918919 0.664732142857143
};
\addplot [semithick, red, forget plot]
table {%
7.25 1.4
10.25 1.4
};
\addplot [semithick, red, forget plot]
table {%
9.5 2.15
10.25 1.4
9.5 0.650000000000001
};
\end{axis}

\end{tikzpicture}

%% file: figs/cut-in.tikz
\begin{tikzpicture}
\definecolor{color0}{rgb}{0.8,1,0.8}
\definecolor{color1}{rgb}{0,0.4375,0.75}

\begin{axis}[
axis background/.style={fill=color0},
height=0.33\figurewidth,
scale only axis,
tick align=outside,
tick pos=left,
ticks=none,
width=\figurewidth,
x grid style={white!69.01960784313725!black},
xmin=-15, xmax=15,
xtick style={color=black},
y grid style={white!69.01960784313725!black},
ymin=-5, ymax=5,
ytick style={color=black}
]
\path [draw=white!80.0!black, fill=white!80.0!black]
(axis cs:-15,2.8)
--(axis cs:15,2.8)
--(axis cs:15,-2.8)
--(axis cs:-15,-2.8)
--cycle;
\addplot [semithick, black]
table {%
-15 2.8
15 2.8
};
\addplot [semithick, black]
table {%
15 -2.8
-15 -2.8
};
\addplot [semithick, white]
table {%
-15 0
-14.5 0
};
\addplot [semithick, white]
table {%
-12.5 0
-11.5 0
};
\addplot [semithick, white]
table {%
-9.5 0
-8.5 0
};
\addplot [semithick, white]
table {%
-6.5 0
-5.5 0
};
\addplot [semithick, white]
table {%
-3.5 0
-2.5 0
};
\addplot [semithick, white]
table {%
-0.499999999999999 0
0.500000000000002 0
};
\addplot [semithick, white]
table {%
2.5 0
3.5 0
};
\addplot [semithick, white]
table {%
5.5 0
6.5 0
};
\addplot [semithick, white]
table {%
8.5 0
9.5 0
};
\addplot [semithick, white]
table {%
11.5 0
12.5 0
};
\addplot [semithick, white]
table {%
14.5 0
15 0
};
\path [draw=black, fill=color1]
(axis cs:-12.25,-1.39598214285714)
--(axis cs:-12.2418918918919,-0.873660714285714)
--(axis cs:-12.1851351351351,-0.672767857142857)
--(axis cs:-12.1040540540541,-0.600446428571429)
--(axis cs:-11.9743243243243,-0.552232142857143)
--(axis cs:-11.5121621621622,-0.495982142857143)
--(axis cs:-8.09864864864865,-0.536160714285714)
--(axis cs:-8.00945945945946,-0.592410714285714)
--(axis cs:-7.87162162162162,-0.769196428571428)
--(axis cs:-7.81486486486487,-0.962053571428571)
--(axis cs:-7.75,-1.39598214285714)
--(axis cs:-7.75,-1.40401785714286)
--(axis cs:-7.81486486486487,-1.83794642857143)
--(axis cs:-7.87162162162162,-2.03080357142857)
--(axis cs:-8.00945945945946,-2.20758928571429)
--(axis cs:-8.09864864864865,-2.26383928571429)
--(axis cs:-11.5121621621622,-2.30401785714286)
--(axis cs:-11.9743243243243,-2.24776785714286)
--(axis cs:-12.1040540540541,-2.19955357142857)
--(axis cs:-12.1851351351351,-2.12723214285714)
--(axis cs:-12.2418918918919,-1.92633928571429)
--(axis cs:-12.25,-1.40401785714286)
--cycle;
\path [draw=black, fill=color1]
(axis cs:-9.33108108108108,-0.568303571428571)
--(axis cs:-9.33918918918919,-0.367410714285714)
--(axis cs:-9.33108108108108,-0.319196428571428)
--(axis cs:-9.29864864864865,-0.343303571428571)
--(axis cs:-9.25,-0.552232142857143)
--cycle;
\path [draw=black, fill=color1]
(axis cs:-9.33108108108108,-2.23169642857143)
--(axis cs:-9.33918918918919,-2.43258928571429)
--(axis cs:-9.33108108108108,-2.48080357142857)
--(axis cs:-9.29864864864865,-2.45669642857143)
--(axis cs:-9.25,-2.24776785714286)
--cycle;
\path [draw=black, fill=white]
(axis cs:-11.7148648648649,-1.39598214285714)
--(axis cs:-11.6986486486486,-1.02633928571429)
--(axis cs:-11.65,-0.801339285714286)
--(axis cs:-11.5851351351351,-0.680803571428571)
--(axis cs:-11.5283783783784,-0.672767857142857)
--(axis cs:-11.0175675675676,-0.809375)
--(axis cs:-11.0662162162162,-0.945982142857143)
--(axis cs:-11.0743243243243,-1.15491071428571)
--(axis cs:-11.0743243243243,-1.64508928571429)
--(axis cs:-11.0662162162162,-1.85401785714286)
--(axis cs:-11.0175675675676,-1.990625)
--(axis cs:-11.5283783783784,-2.12723214285714)
--(axis cs:-11.5851351351351,-2.11919642857143)
--(axis cs:-11.65,-1.99866071428571)
--(axis cs:-11.6986486486486,-1.77366071428571)
--(axis cs:-11.7148648648649,-1.40401785714286)
--cycle;
\path [draw=black, fill=white]
(axis cs:-8.91756756756757,-1.39598214285714)
--(axis cs:-8.94189189189189,-0.962053571428571)
--(axis cs:-9.03108108108108,-0.688839285714286)
--(axis cs:-9.08783783783784,-0.632589285714286)
--(axis cs:-9.61486486486486,-0.841517857142857)
--(axis cs:-9.56621621621622,-1.034375)
--(axis cs:-9.55,-1.203125)
--(axis cs:-9.55,-1.596875)
--(axis cs:-9.56621621621622,-1.765625)
--(axis cs:-9.61486486486486,-1.95848214285714)
--(axis cs:-9.08783783783784,-2.16741071428571)
--(axis cs:-9.03108108108108,-2.11116071428571)
--(axis cs:-8.94189189189189,-1.83794642857143)
--(axis cs:-8.91756756756757,-1.40401785714286)
--cycle;
\path [draw=black, fill=white]
(axis cs:-11.0256756756757,-0.568303571428571)
--(axis cs:-10.8148648648649,-0.568303571428571)
--(axis cs:-10.8148648648649,-0.720982142857143)
--(axis cs:-10.9202702702703,-0.672767857142857)
--cycle;
\path [draw=black, fill=white]
(axis cs:-11.0256756756757,-2.23169642857143)
--(axis cs:-10.8148648648649,-2.23169642857143)
--(axis cs:-10.8148648648649,-2.07901785714286)
--(axis cs:-10.9202702702703,-2.12723214285714)
--cycle;
\path [draw=black, fill=white]
(axis cs:-10.7662162162162,-0.737053571428571)
--(axis cs:-10.7662162162162,-0.608482142857143)
--(axis cs:-10.7337837837838,-0.576339285714286)
--(axis cs:-10.2067567567568,-0.576339285714286)
--(axis cs:-10.1824324324324,-0.616517857142857)
--(axis cs:-10.2310810810811,-0.761160714285714)
--(axis cs:-10.3040540540541,-0.793303571428571)
--(axis cs:-10.5716216216216,-0.777232142857143)
--cycle;
\path [draw=black, fill=white]
(axis cs:-10.7662162162162,-2.06294642857143)
--(axis cs:-10.7662162162162,-2.19151785714286)
--(axis cs:-10.7337837837838,-2.22366071428571)
--(axis cs:-10.2067567567568,-2.22366071428571)
--(axis cs:-10.1824324324324,-2.18348214285714)
--(axis cs:-10.2310810810811,-2.03883928571429)
--(axis cs:-10.3040540540541,-2.00669642857143)
--(axis cs:-10.5716216216216,-2.02276785714286)
--cycle;
\path [draw=black, fill=white]
(axis cs:-10.15,-0.817410714285714)
--(axis cs:-10.0202702702703,-0.568303571428571)
--(axis cs:-9.28243243243243,-0.568303571428571)
--(axis cs:-9.29054054054054,-0.616517857142857)
--(axis cs:-9.70405405405405,-0.793303571428571)
--cycle;
\path [draw=black, fill=white]
(axis cs:-10.15,-1.98258928571429)
--(axis cs:-10.0202702702703,-2.23169642857143)
--(axis cs:-9.28243243243243,-2.23169642857143)
--(axis cs:-9.29054054054054,-2.18348214285714)
--(axis cs:-9.70405405405405,-2.00669642857143)
--cycle;
\path [draw=black, fill=white]
(axis cs:-8.2527027027027,-0.552232142857143)
--(axis cs:-8.09054054054054,-0.560267857142857)
--(axis cs:-7.96891891891892,-0.680803571428571)
--(axis cs:-7.91216216216216,-0.777232142857143)
--(axis cs:-7.89594594594595,-0.873660714285714)
--(axis cs:-7.89594594594595,-1.04241071428571)
--(axis cs:-7.98513513513513,-0.889732142857143)
--cycle;
\path [draw=black, fill=white]
(axis cs:-8.2527027027027,-2.24776785714286)
--(axis cs:-8.09054054054054,-2.23973214285714)
--(axis cs:-7.96891891891892,-2.11919642857143)
--(axis cs:-7.91216216216216,-2.02276785714286)
--(axis cs:-7.89594594594595,-1.92633928571429)
--(axis cs:-7.89594594594595,-1.75758928571429)
--(axis cs:-7.98513513513513,-1.91026785714286)
--cycle;
\path [draw=black, fill=red]
(axis cs:-7.25,1.40401785714286)
--(axis cs:-7.24189189189189,1.92633928571429)
--(axis cs:-7.18513513513513,2.12723214285714)
--(axis cs:-7.10405405405405,2.19955357142857)
--(axis cs:-6.97432432432432,2.24776785714286)
--(axis cs:-6.51216216216216,2.30401785714286)
--(axis cs:-3.09864864864865,2.26383928571429)
--(axis cs:-3.00945945945946,2.20758928571429)
--(axis cs:-2.87162162162162,2.03080357142857)
--(axis cs:-2.81486486486487,1.83794642857143)
--(axis cs:-2.75,1.40401785714286)
--(axis cs:-2.75,1.39598214285714)
--(axis cs:-2.81486486486487,0.962053571428571)
--(axis cs:-2.87162162162162,0.769196428571429)
--(axis cs:-3.00945945945946,0.592410714285714)
--(axis cs:-3.09864864864865,0.536160714285714)
--(axis cs:-6.51216216216216,0.495982142857143)
--(axis cs:-6.97432432432432,0.552232142857143)
--(axis cs:-7.10405405405405,0.600446428571428)
--(axis cs:-7.18513513513513,0.672767857142857)
--(axis cs:-7.24189189189189,0.873660714285714)
--(axis cs:-7.25,1.39598214285714)
--cycle;
\path [draw=black, fill=red]
(axis cs:-4.33108108108108,2.23169642857143)
--(axis cs:-4.33918918918919,2.43258928571429)
--(axis cs:-4.33108108108108,2.48080357142857)
--(axis cs:-4.29864864864865,2.45669642857143)
--(axis cs:-4.25,2.24776785714286)
--cycle;
\path [draw=black, fill=red]
(axis cs:-4.33108108108108,0.568303571428571)
--(axis cs:-4.33918918918919,0.367410714285714)
--(axis cs:-4.33108108108108,0.319196428571428)
--(axis cs:-4.29864864864865,0.343303571428571)
--(axis cs:-4.25,0.552232142857143)
--cycle;
\path [draw=black, fill=white]
(axis cs:-6.71486486486486,1.40401785714286)
--(axis cs:-6.69864864864865,1.77366071428571)
--(axis cs:-6.65,1.99866071428571)
--(axis cs:-6.58513513513514,2.11919642857143)
--(axis cs:-6.52837837837838,2.12723214285714)
--(axis cs:-6.01756756756757,1.990625)
--(axis cs:-6.06621621621622,1.85401785714286)
--(axis cs:-6.07432432432432,1.64508928571429)
--(axis cs:-6.07432432432432,1.15491071428571)
--(axis cs:-6.06621621621622,0.945982142857143)
--(axis cs:-6.01756756756757,0.809375)
--(axis cs:-6.52837837837838,0.672767857142857)
--(axis cs:-6.58513513513514,0.680803571428571)
--(axis cs:-6.65,0.801339285714285)
--(axis cs:-6.69864864864865,1.02633928571429)
--(axis cs:-6.71486486486486,1.39598214285714)
--cycle;
\path [draw=black, fill=white]
(axis cs:-3.91756756756757,1.40401785714286)
--(axis cs:-3.94189189189189,1.83794642857143)
--(axis cs:-4.03108108108108,2.11116071428571)
--(axis cs:-4.08783783783784,2.16741071428571)
--(axis cs:-4.61486486486486,1.95848214285714)
--(axis cs:-4.56621621621622,1.765625)
--(axis cs:-4.55,1.596875)
--(axis cs:-4.55,1.203125)
--(axis cs:-4.56621621621622,1.034375)
--(axis cs:-4.61486486486486,0.841517857142857)
--(axis cs:-4.08783783783784,0.632589285714286)
--(axis cs:-4.03108108108108,0.688839285714286)
--(axis cs:-3.94189189189189,0.962053571428571)
--(axis cs:-3.91756756756757,1.39598214285714)
--cycle;
\path [draw=black, fill=white]
(axis cs:-6.02567567567568,2.23169642857143)
--(axis cs:-5.81486486486487,2.23169642857143)
--(axis cs:-5.81486486486487,2.07901785714286)
--(axis cs:-5.92027027027027,2.12723214285714)
--cycle;
\path [draw=black, fill=white]
(axis cs:-6.02567567567568,0.568303571428571)
--(axis cs:-5.81486486486487,0.568303571428571)
--(axis cs:-5.81486486486487,0.720982142857143)
--(axis cs:-5.92027027027027,0.672767857142857)
--cycle;
\path [draw=black, fill=white]
(axis cs:-5.76621621621622,2.06294642857143)
--(axis cs:-5.76621621621622,2.19151785714286)
--(axis cs:-5.73378378378378,2.22366071428571)
--(axis cs:-5.20675675675676,2.22366071428571)
--(axis cs:-5.18243243243243,2.18348214285714)
--(axis cs:-5.23108108108108,2.03883928571429)
--(axis cs:-5.30405405405405,2.00669642857143)
--(axis cs:-5.57162162162162,2.02276785714286)
--cycle;
\path [draw=black, fill=white]
(axis cs:-5.76621621621622,0.737053571428571)
--(axis cs:-5.76621621621622,0.608482142857143)
--(axis cs:-5.73378378378378,0.576339285714286)
--(axis cs:-5.20675675675676,0.576339285714286)
--(axis cs:-5.18243243243243,0.616517857142857)
--(axis cs:-5.23108108108108,0.761160714285714)
--(axis cs:-5.30405405405405,0.793303571428571)
--(axis cs:-5.57162162162162,0.777232142857143)
--cycle;
\path [draw=black, fill=white]
(axis cs:-5.15,1.98258928571429)
--(axis cs:-5.02027027027027,2.23169642857143)
--(axis cs:-4.28243243243243,2.23169642857143)
--(axis cs:-4.29054054054054,2.18348214285714)
--(axis cs:-4.70405405405405,2.00669642857143)
--cycle;
\path [draw=black, fill=white]
(axis cs:-5.15,0.817410714285714)
--(axis cs:-5.02027027027027,0.568303571428571)
--(axis cs:-4.28243243243243,0.568303571428571)
--(axis cs:-4.29054054054054,0.616517857142857)
--(axis cs:-4.70405405405405,0.793303571428571)
--cycle;
\path [draw=black, fill=white]
(axis cs:-3.2527027027027,2.24776785714286)
--(axis cs:-3.09054054054054,2.23973214285714)
--(axis cs:-2.96891891891892,2.11919642857143)
--(axis cs:-2.91216216216216,2.02276785714286)
--(axis cs:-2.89594594594595,1.92633928571429)
--(axis cs:-2.89594594594595,1.75758928571429)
--(axis cs:-2.98513513513514,1.91026785714286)
--cycle;
\path [draw=black, fill=white]
(axis cs:-3.2527027027027,0.552232142857143)
--(axis cs:-3.09054054054054,0.560267857142857)
--(axis cs:-2.96891891891892,0.680803571428571)
--(axis cs:-2.91216216216216,0.777232142857143)
--(axis cs:-2.89594594594595,0.873660714285714)
--(axis cs:-2.89594594594595,1.04241071428571)
--(axis cs:-2.98513513513514,0.889732142857143)
--cycle;
\addplot [black]
table {%
-11.6337837837838 -0.656696428571429
-11.8608108108108 -0.664732142857143
-12.0716216216216 -0.720982142857143
-12.1364864864865 -0.801339285714286
-12.1364864864865 -1.99866071428571
-12.0716216216216 -2.07901785714286
-11.8608108108108 -2.13526785714286
-11.6337837837838 -2.14330357142857
};
\addplot [black]
table {%
-8.00945945945946 -0.970089285714286
-7.98513513513513 -0.945982142857143
-7.89594594594595 -1.08258928571429
-7.89594594594595 -1.71741071428571
-7.98513513513513 -1.85401785714286
-8.00945945945946 -1.82991071428571
};
\addplot [black]
table {%
-8.26081081081081 -0.664732142857143
-9.0472972972973 -0.632589285714286
-8.00945945945946 -0.970089285714286
-8.00945945945946 -1.82991071428571
-9.0472972972973 -2.16741071428571
-8.26081081081081 -2.13526785714286
};
\addplot [semithick, red]
table {%
-7.75 -1.4
-2.75 -1.4
};
\addplot [semithick, red]
table {%
-3.5 -0.649999999999999
-2.75 -1.4
-3.5 -2.15
};
\addplot [black]
table {%
-6.63378378378378 2.14330357142857
-6.86081081081081 2.13526785714286
-7.07162162162162 2.07901785714286
-7.13648648648649 1.99866071428571
-7.13648648648649 0.801339285714285
-7.07162162162162 0.720982142857143
-6.86081081081081 0.664732142857143
-6.63378378378378 0.656696428571428
};
\addplot [black]
table {%
-3.00945945945946 1.82991071428571
-2.98513513513514 1.85401785714286
-2.89594594594595 1.71741071428571
-2.89594594594595 1.08258928571429
-2.98513513513514 0.945982142857143
-3.00945945945946 0.970089285714286
};
\addplot [black]
table {%
-3.26081081081081 2.13526785714286
-4.0472972972973 2.16741071428571
-3.00945945945946 1.82991071428571
-3.00945945945946 0.970089285714286
-4.0472972972973 0.632589285714286
-3.26081081081081 0.664732142857143
};
\addplot [semithick, red]
table {%
-2.75 1.4
-0.75 1.4
-0.434210526315789 1.38090582476381
-0.118421052631579 1.32414413838089
0.197368421052632 1.23126325168908
0.513157894736842 1.10479671315495
0.828947368421053 0.948194200276037
1.14473684210526 0.765727421371397
1.46052631578947 0.562373594514157
1.77631578947368 0.343679681997119
2.09210526315789 0.115611083661265
2.40789473684211 -0.115611083661265
2.72368421052632 -0.343679681997119
3.03947368421053 -0.562373594514157
3.35526315789474 -0.765727421371398
3.67105263157895 -0.948194200276037
3.98684210526316 -1.10479671315495
4.30263157894737 -1.23126325168908
4.61842105263158 -1.32414413838089
4.93421052631579 -1.38090582476381
5.25 -1.4
7.25 -1.4
};
\addplot [semithick, red]
table {%
6.5 -0.65
7.25 -1.4
6.5 -2.15
};
\end{axis}

\end{tikzpicture}

%% file: figs/singular_vectors.tikz
\begin{tikzpicture}

\begin{axis}[
height=\figureheight,
legend cell align={left},
legend style={
  fill opacity=0.8,
  draw opacity=1,
  text opacity=1,
  at={(0.5,0.55)},
  anchor=north west,
  draw=white!80!black
},
legend style={nodes={scale=0.7, transform shape}},
scaled y ticks=false,
tick align=outside,
tick pos=left,
width=\figurewidth,
x grid style={white!69.0196078431373!black},
xlabel={Coordinate},
xmin=-1.45, xmax=52.45,
xtick style={color=black},
xticklabel style={align=center},
y grid style={white!69.0196078431373!black},
ylabel={Acceleration [\si{\meter\per\second\squared}]},
ymin=-0.913690682752387, ymax=0.0888827608467326,
ytick style={color=black},
yticklabel style={/pgf/number format/fixed,/pgf/number format/precision=3}
]
\addplot [semithick, black, mark=*, mark size=1.5, mark options={solid}, only marks]
table {%
1 -0.435540945521222
2 -0.484456239278613
3 -0.528504116878438
4 -0.567266973765254
5 -0.600583456258151
6 -0.628368172909993
7 -0.651051491702439
8 -0.669579148013173
9 -0.685479629733573
10 -0.700474036723259
11 -0.716253418641113
12 -0.731443602672833
13 -0.744793586276749
14 -0.756360111374774
15 -0.766526038447426
16 -0.775417694262796
17 -0.783230834333785
18 -0.790207564842927
19 -0.796498410967959
20 -0.801817462106627
21 -0.805621112221214
22 -0.808608495461709
23 -0.811115036251276
24 -0.813291203402015
25 -0.815276534253683
26 -0.817154759282594
27 -0.819006483831168
28 -0.820830320612151
29 -0.82267005029741
30 -0.824417399685458
31 -0.826308372999131
32 -0.828076123164863
33 -0.829866082883162
34 -0.831733985400506
35 -0.83368158339319
36 -0.835772333851804
37 -0.837836925743977
38 -0.839670622274511
39 -0.840991354397097
40 -0.842065280644667
41 -0.844205341809974
42 -0.847673213912877
43 -0.851373695706113
44 -0.854691149369739
45 -0.857158685711576
46 -0.858592513687337
47 -0.858935842219001
48 -0.858077678539475
49 -0.855928302478624
50 -0.852478224471344
};
\addlegendentry{$\svdmean\elementdivision\weights$}
\addplot [semithick, black]
table {%
1 -0.157310030024398
2 -0.158968203020566
3 -0.160678588664791
4 -0.162419985183483
5 -0.164153556540662
6 -0.165782280536184
7 -0.167444086741668
8 -0.169341209613531
9 -0.171342991094525
10 -0.172964976700256
11 -0.173622367941652
12 -0.173864225973941
13 -0.174209759130503
14 -0.174727020035538
15 -0.175397745057976
16 -0.176177111686744
17 -0.177127030603692
18 -0.178309879031586
19 -0.179621570335408
20 -0.181015733185071
21 -0.182669840296596
22 -0.184354128460469
23 -0.185968876284575
24 -0.187466517374281
25 -0.188704901811098
26 -0.189581275316016
27 -0.190227706205667
28 -0.190674017747948
29 -0.19086010013913
30 -0.190881102366686
31 -0.19075795793159
32 -0.190529012514308
33 -0.190120364402179
34 -0.189529726770638
35 -0.188783597622226
36 -0.187932845590168
37 -0.18696718604271
38 -0.185917314205932
39 -0.184799799784037
40 -0.18346064325966
41 -0.181340131982021
42 -0.178081100019484
43 -0.173771760511905
44 -0.168386498394351
45 -0.162194864136722
46 -0.15527130360012
47 -0.147650069367888
48 -0.139429121014092
49 -0.130612430994943
50 -0.121196106329894
};
\addlegendentry{$\svduvec{1}\elementdivision\weights$}
\addplot [semithick, gray]
table {%
1 -0.00136282835632667
2 -0.0420032659430694
3 -0.0810860840765591
4 -0.118071674650332
5 -0.15285163833347
6 -0.185261727395398
7 -0.215422463573454
8 -0.243789753570802
9 -0.270305105918527
10 -0.294292993837814
11 -0.314608521839236
12 -0.331295809354394
13 -0.344957304176465
14 -0.355813961900325
15 -0.364072741210934
16 -0.369916899474496
17 -0.373552931337142
18 -0.375194317603775
19 -0.375167561975016
20 -0.373583632858443
21 -0.37081909142446
22 -0.367075988319313
23 -0.362278972716367
24 -0.356279661259095
25 -0.349368833601081
26 -0.341677189195124
27 -0.333319209280436
28 -0.324284534407778
29 -0.314586371718226
30 -0.304294404376288
31 -0.293654563315591
32 -0.282947465121744
33 -0.272269743014781
34 -0.261668568160111
35 -0.251208813790001
36 -0.240886822240635
37 -0.230482023258236
38 -0.220028325493067
39 -0.209084868321627
40 -0.197292510727745
41 -0.184985696566025
42 -0.172340084427513
43 -0.159987171691079
44 -0.148090339226624
45 -0.137093517863201
46 -0.127193664602323
47 -0.118491479801131
48 -0.111396917950504
49 -0.106115361687174
50 -0.102690320466503
};
\addlegendentry{$\svduvec{2}\elementdivision\weights$}
\addplot [semithick, black, dashed]
table {%
1 0.0433112406831362
2 0.0178339779911003
3 -0.00561639728041875
4 -0.0266751435460045
5 -0.0451455312506936
6 -0.0607455930135511
7 -0.073506894713016
8 -0.0833363274462632
9 -0.0905602341562927
10 -0.0959362673211083
11 -0.100317979806822
12 -0.103871204681654
13 -0.106281919255135
14 -0.107599565766502
15 -0.108161560466658
16 -0.108028985975503
17 -0.107383767155108
18 -0.10629580483806
19 -0.10520570313042
20 -0.104220589834994
21 -0.103252382824176
22 -0.102090425320024
23 -0.100640333669298
24 -0.099054182584518
25 -0.0974839693475158
26 -0.0959854368985565
27 -0.0946310709127666
28 -0.0933136614651436
29 -0.0918917818359289
30 -0.0901888002951871
31 -0.088526167824415
32 -0.0878551449182638
33 -0.0881734597694951
34 -0.0893221811533244
35 -0.0912953615028099
36 -0.093993504049254
37 -0.0971858320940808
38 -0.100668232717303
39 -0.104215881491637
40 -0.108463687362682
41 -0.113784256110131
42 -0.119428214159054
43 -0.1252249853517
44 -0.131084938318737
45 -0.136726202472846
46 -0.141918332082748
47 -0.146567342696991
48 -0.15046652486832
49 -0.153582511394659
50 -0.155890703171959
};
\addlegendentry{$\svduvec{3}\elementdivision\weights$}
\addplot [semithick, gray, dashed]
table {%
1 -0.232977578274069
2 -0.267978504296498
3 -0.304348939306144
4 -0.341407237229486
5 -0.378824952320792
6 -0.416338054236158
7 -0.453990244325513
8 -0.492454941989693
9 -0.531530793603985
10 -0.570801577698131
11 -0.609742094212009
12 -0.646955093920054
13 -0.681643706441162
14 -0.713678849864519
15 -0.743011691309451
16 -0.769482289400382
17 -0.792902284599826
18 -0.813263058840648
19 -0.830487607770964
20 -0.844117176156498
21 -0.85394416140142
22 -0.860769680171013
23 -0.86523628801555
24 -0.867576168480605
25 -0.868119162588791
26 -0.867067120520703
27 -0.864516384577834
28 -0.86061933499108
29 -0.855817250947285
30 -0.850762903490817
31 -0.84661348744096
32 -0.843398395221015
33 -0.841284472848189
34 -0.840043702911047
35 -0.839731747421931
36 -0.840521406926975
37 -0.842008339099686
38 -0.844129666007063
39 -0.845792219160875
40 -0.844997408171445
41 -0.840162800391258
42 -0.830297226451607
43 -0.81447763993109
44 -0.792676287969438
45 -0.764550864935338
46 -0.729896444185154
47 -0.688677958762397
48 -0.640670119829562
49 -0.585797025861995
50 -0.524041903379691
};
\addlegendentry{$\svduvec{4}\elementdivision\weights$}
\end{axis}

\end{tikzpicture}

%% file: figs/corner_cases.tikz
\begin{tikzpicture}

\begin{axis}[
height=\figureheight,
legend cell align={left},
legend style={fill opacity=0.8, draw opacity=1, text opacity=1, draw=white!80!black},
legend style={nodes={scale=0.7, transform shape}},
scaled y ticks=false,
tick align=outside,
tick pos=left,
width=\figurewidth,
x grid style={white!69.0196078431373!black},
xlabel={Time [s]},
xmin=-0.541897397398134, xmax=11.3798453453608,
xtick style={color=black},
xticklabel style={align=center},
y grid style={white!69.0196078431373!black},
ylabel={Speed [km/h]},
ymin=1.47866370760589, ymax=88.9542351697775,
ytick style={color=black},
yticklabel style={/pgf/number format/fixed,/pgf/number format/precision=3}
]
\addplot [semithick, black]
table {%
0 43.1543584813747
0.0706122448977929 42.6331459309908
0.141224489795586 42.0762590899685
0.211836734693379 41.48790260599
0.282448979591171 40.8722811267221
0.353061224488964 40.2335992998314
0.423673469386757 39.5760617729777
0.49428571428455 38.9038731938364
0.564897959182343 38.221238210074
0.635510204080136 37.5323614693573
0.706122448977929 36.8413646069611
0.776734693875722 36.1496298492335
0.847346938773514 35.4572112242533
0.917959183671307 34.7641627600992
0.9885714285691 34.0705384848498
1.05918367346689 33.3763924265842
1.12979591836469 32.6817786133812
1.20040816326248 31.9867510733193
1.27102040816027 31.2913638344776
1.34163265305806 30.5956709249348
1.41224489795586 29.8992707462385
1.48285714285365 29.1975471545164
1.55346938775144 28.4848588462047
1.62408163264924 27.7555645177385
1.69469387754703 27.0040228655528
1.76530612244482 26.22459258607
1.83591836734261 25.4116323757356
1.90653061224041 24.5595009309846
1.9771428571382 23.6625569482521
2.04775510203599 22.7151591239512
2.11836734693379 21.7154979177605
2.18897959183158 20.6775165936987
2.25959183672937 19.616861421661
2.33020408162716 18.5491786715427
2.40081632652496 17.4901146132448
2.47142857142275 16.4553155166623
2.54204081632054 15.4604276517099
2.61265306121834 14.5210972882694
2.68326530611613 13.6529706962357
2.75387755101392 12.8716941455039
2.82448979591172 12.1858607812157
2.89510204080951 11.5895166785429
2.9657142857073 11.0762670923982
3.03632653060509 10.6397172776941
3.10693877550289 10.2734724893706
3.17755102040068 9.97113798230997
3.24816326529847 9.72631901142469
3.31877551019626 9.53262083162723
3.38938775509406 9.38364869784719
3.45999999999185 9.27300786497733
};
\addlegendentry{Measured}
\addplot [semithick, gray]
table {%
0 42.8718945173594
0.0673357025923014 42.5260898586303
0.134671405184603 42.1469477723515
0.202007107776904 41.7355119410761
0.269342810369205 41.2930644648874
0.336678512961507 40.8211084308757
0.404014215553808 40.3209373504581
0.47134991814611 39.7931210844232
0.538685620738411 39.2380404722459
0.606021323330712 38.6561901153024
0.673357025923014 38.0483313238389
0.740692728515315 37.4159936257509
0.808028431107616 36.7611029154193
0.875364133699918 36.0855865750516
0.942699836292219 35.3913014037158
1.01003553888452 34.6801681526276
1.07737124147682 33.9541115345852
1.14470694406912 33.2149535112425
1.21204264666142 32.4645542816755
1.27937834925373 31.7051006480701
1.34671405184603 30.9388967388995
1.41404975443833 30.1677648833195
1.48138545703063 29.3931682426946
1.54872115962293 28.6164493013986
1.61605686221523 27.8387998108297
1.68339256480753 27.0613313441332
1.75072826739984 26.2850069779082
1.81806396999214 25.5106945650836
1.88539967258444 24.7390508113591
1.95273537517674 23.9703601289733
2.02007107776904 23.2042171623629
2.08740678036134 22.4402306772659
2.15474248295364 21.6779193256029
2.22207818554595 20.9169043056522
2.28941388813825 20.1567583131572
2.35674959073055 19.3969207230781
2.42408529332285 18.6371037042319
2.49142099591515 17.877081785269
2.55875669850745 17.1172917722543
2.62609240109975 16.3592640085931
2.69342810369205 15.6053154002716
2.76076380628436 14.8583913606292
2.82809950887666 14.1220911145334
2.89543521146896 13.4001788045932
2.96277091406126 12.696532394014
3.03010661665356 12.0151259243917
3.09744231924586 11.3599392688281
3.16477802183816 10.7350133427275
3.23211372443047 10.1444331273293
3.29944942702277 9.59229341810905
};
\addlegendentry{Approximation}
\addplot [semithick, black, forget plot]
table {%
0 80.5901333797399
0.0926530612244705 79.5652278340673
0.185306122448941 78.513360679651
0.277959183673411 77.4393899574736
0.370612244897882 76.3481737085182
0.463265306122352 75.2445699737682
0.555918367346823 74.1334367942062
0.648571428571293 73.0196322108155
0.741224489795764 71.9080142645791
0.833877551020234 70.8034409964799
0.926530612244705 69.7107704475013
1.01918367346918 68.634860658626
1.11183673469365 67.5805696708372
1.20448979591812 66.5527555251181
1.29714285714259 65.5562762624511
1.38979591836706 64.5959899238201
1.48244897959153 63.676754550208
1.575102040816 62.803428182597
1.66775510204047 61.980868861971
1.76040816326494 61.2139346293133
1.85306122448941 60.5074835256056
1.94571428571388 59.8663735918323
2.03836734693835 59.2954628689763
2.13102040816282 58.7996093980194
2.22367346938729 58.3805142400786
2.31632653061176 58.0295816914604
2.40897959183623 57.7374194739057
2.5016326530607 57.4946353091544
2.59428571428517 57.2918369189485
2.68693877550964 57.1196320250288
2.77959183673411 56.9686283491358
2.87224489795859 56.8294336130109
2.96489795918306 56.6926555383949
3.05755102040753 56.5489018470289
3.150204081632 56.3887802606537
3.24285714285647 56.2028985010101
3.33551020408094 55.9818642898397
3.42816326530541 55.7162853488829
3.52081632652988 55.3967693998804
3.61346938775435 55.0139241645744
3.70612244897882 54.5583573647046
3.79877551020329 54.0206767220119
3.89142857142776 53.391489958239
3.98408163265223 52.661404795125
4.0767346938767 51.8210289544107
4.16938775510117 50.8609701578396
4.26204081632564 49.77183612715
4.35469387755011 48.5442345840829
4.44734693877458 47.1687732503822
4.53999999999905 45.6360598477858
};
\addplot [semithick, gray, forget plot]
table {%
0 80.6962388225165
0.0886359165463777 80.4092914746164
0.177271833092755 80.0843138585799
0.265907749639133 79.7228072149026
0.354543666185511 79.3264820360247
0.443179582731889 78.8971975324759
0.531815499278266 78.4366415765591
0.620451415824644 77.9457588010891
0.709087332371022 77.4251568115626
0.7977232489174 76.8753322584595
0.886359165463777 76.2968529479466
0.974995082010155 75.6914844081213
1.06363099855653 75.0616438494712
1.15226691510291 74.4097548227776
1.24090283164929 73.7381409617784
1.32953874819567 73.0491262332025
1.41817466474204 72.3450484663164
1.50681058128842 71.6281368725213
1.5954464978348 70.9005876550083
1.68408241438118 70.1649019666014
1.77271833092755 69.4237701429294
1.86135424747393 68.679160008236
1.94999016402031 67.9326073150777
2.03862608056669 67.1855377185461
2.12726199711307 66.4391074415581
2.21589791365944 65.6943128871599
2.30453383020582 64.9520713815277
2.3931697467522 64.2132141282133
2.48180566329858 63.4783090392185
2.57044157984495 62.747598720569
2.65907749639133 62.0206627841459
2.74771341293771 61.2972262221112
2.83634932948409 60.5768542523317
2.92498524603047 59.8591732731446
3.01362116257684 59.1437672081234
3.10225707912322 58.430104353747
3.1908929956696 57.717927214041
3.27952891221598 57.0070397019724
3.36816482876235 56.2979558517454
3.45680074530873 55.5922675336689
3.54543666185511 54.8918616262492
3.63407257840149 54.1988525905225
3.72270849494786 53.5159956858341
3.81134441149424 52.8462017588145
3.89998032804062 52.1925771514775
3.988616244587 51.55831456204
4.07725216113338 50.9466073488384
4.16588807767975 50.3606764243822
4.25452399422613 49.803755223549
4.34315991077251 49.2790741621519
};
\addplot [semithick, black, forget plot]
table {%
0 56.0271873536443
0.121428571428809 55.9198770962547
0.242857142857618 55.7114649055921
0.364285714286427 55.4072116529893
0.485714285715236 55.0123782098041
0.607142857144045 54.5322254474138
0.728571428572854 53.9720142371552
0.850000000001663 53.3370054503859
0.971428571430472 52.6324599584635
1.09285714285928 51.8636386327606
1.21428571428809 51.0367524521841
1.3357142857169 50.1615824588195
1.45714285714571 49.2482511933519
1.57857142857452 48.306881196448
1.70000000000333 47.3475950087849
1.82142857143214 46.3805151710404
1.94285714286094 45.4157642238884
2.06428571428975 44.4634647080061
2.18571428571856 43.5337391640639
2.30714285714737 42.6367101327431
2.42857142857618 41.7818726477798
2.55000000000499 40.9663209750631
2.6714285714338 40.1823312803029
2.79285714286261 39.422179729224
2.91428571429142 38.6781424875429
3.03571428572023 37.9424957209749
3.15714285714903 37.2075155952394
3.27857142857784 36.4654782760529
3.40000000000665 35.7086599291362
3.52142857143546 34.9293367202031
3.64285714286427 34.1197848149702
3.76428571429308 33.2745444994711
3.88571428572189 32.3913936545837
4.0071428571507 31.4681350026223
4.12857142857951 30.5025712659105
4.25000000000832 29.4925051667824
4.37142857143712 28.435739427551
4.49285714286593 27.3300767705396
4.61428571429474 26.1733199180841
4.73571428572355 24.9632715924959
4.85714285715236 23.6977345160982
4.97857142858117 22.3821055644034
5.10000000000998 21.0600023133743
5.22142857143879 19.7803275273103
5.3428571428676 18.5919839705107
5.46428571429641 17.5438744072357
5.58571428572521 16.6849016018123
5.70714285715402 16.0639683185399
5.82857142858283 15.7299773216435
5.95000000001164 15.7318313754852
};
\addplot [semithick, gray, forget plot]
table {%
0 55.9415195867796
0.122386009308822 55.50551528444
0.244772018617644 55.0331944600153
0.367158027926466 54.5257709262305
0.489544037235288 53.9847143507734
0.61193004654411 53.4116678853293
0.734316055852932 52.8080358012088
0.856702065161754 52.1743232906207
0.979088074470575 51.5105881835959
1.1014740837794 50.8166182876504
1.22386009308822 50.0920187622195
1.34624610239704 49.3378288921889
1.46863211170586 48.5560344349805
1.59101812101468 47.7486947904742
1.71340413032351 46.9178208443089
1.83579013963233 46.0655161930862
1.95817614894115 45.1939617521803
2.08056215824997 44.3052539140914
2.20294816755879 43.4016154208946
2.32533417686762 42.4857468189907
2.44772018617644 41.560699272666
2.57010619548526 40.6287177540991
2.69249220479408 39.6914687824226
2.8148782141029 38.750423401234
2.93726422341173 37.8068628686054
3.05965023272055 36.8619458756145
3.18203624202937 35.9167195191894
3.30442225133819 34.9720998997363
3.42680826064701 34.0286776292836
3.54919426995584 33.0865994272171
3.67158027926466 32.14520608932
3.79396628857348 31.2041296721563
3.9163522978823 30.2628542859871
4.03873830719112 29.3209684607947
4.16112431649995 28.3780203349347
4.28351032580877 27.4333811436014
4.40589633511759 26.4867080433424
4.52828234442641 25.5377476174918
4.65066835373523 24.5870191825261
4.77305436304406 23.6363968889667
4.89544037235288 22.6884934328401
5.0178263816617 21.7464464362469
5.14021239097052 20.8145015607925
5.26259840027934 19.8971999153359
5.38498440958817 18.9994356072418
5.50737041889699 18.1262772709277
5.62975642820581 17.2828249133151
5.75214243751463 16.4743528320881
5.87452844682345 15.7062379809037
5.99691445613227 14.9838733367842
};
\addplot [semithick, black, forget plot]
table {%
0 60.6529069005823
0.0887755102042004 60.015028131219
0.177551020408401 59.3322497327286
0.266326530612601 58.6088481292328
0.355102040816802 57.8490997448531
0.443877551021002 57.0572810037011
0.532653061225203 56.2376683299095
0.621428571429403 55.3945381476
0.710204081633603 54.532166880894
0.798979591837804 53.6548309539131
0.887755102042004 52.7668067907787
0.976530612246205 51.8723708156125
1.06530612245041 50.9757994525358
1.15408163265461 50.0807305390766
1.24285714285881 49.1880466886812
1.33163265306301 48.298313530439
1.42040816326721 47.4120966934371
1.50918367347141 46.5299618067638
1.59795918367561 45.6524744995072
1.68673469387981 44.7802004007555
1.77551020408401 43.9137051395967
1.86428571428821 43.0535543451191
1.95306122449241 42.2003136464108
2.04183673469661 41.35454867256
2.13061224490081 40.5168250526548
2.21938775510501 39.6877084157863
2.30816326530921 38.8677643910399
2.39693877551341 38.0575586075038
2.48571428571761 37.2576566942662
2.57448979592181 36.4686242804151
2.66326530612601 35.6910269950387
2.75204081633021 34.9254304672252
2.84081632653441 34.1724003260628
2.92959183673861 33.4325022006396
3.01836734694281 32.7063017200438
3.10714285714701 31.9943645133682
3.19591836735122 31.2972562096965
3.28469387755542 30.6155424381168
3.37346938775962 29.9497888277173
3.46224489796382 29.3003521096746
3.55102040816802 28.6668847226509
3.63979591837222 28.0489817082876
3.72857142857642 27.4462381082263
3.81734693878062 26.8582489641087
3.90612244898482 26.2846093175764
3.99489795918902 25.7249142102755
4.08367346939322 25.1787586838428
4.17244897959742 24.6457377799201
4.26122448980162 24.125446540149
4.35000000000582 23.6174800061709
};
\addplot [semithick, gray, forget plot]
table {%
0 60.7996868355175
0.0847326402592244 60.434330261495
0.169465280518449 60.0294316344249
0.254197920777673 59.5868295407036
0.338930561036897 59.1086386135718
0.423663201296122 58.597244712764
0.508395841555346 58.0547925785318
0.593128481814571 57.4826783618145
0.677861122073795 56.8819479893812
0.762593762333019 56.2535038759202
0.847326402592244 55.5982404111227
0.932059042851468 54.9178399016512
1.01679168311069 54.2144952945835
1.10152432336992 53.4903844513906
1.18625696362914 52.7475267517744
1.27098960388837 51.9879814175375
1.35572224414759 51.2137604834544
1.44045488440681 50.4267445298772
1.52518752466604 49.628758710442
1.60992016492526 48.8219402597146
1.69465280518449 48.0085877312064
1.77938544544371 47.190532888806
1.86411808570294 46.3692617901262
1.94885072596216 45.5461090317401
2.03358336622138 44.7222293518911
2.11831600648061 43.8986865313744
2.20304864673983 43.0763801193277
2.28778128699906 42.2561384014607
2.37251392725828 41.4386133236093
2.45724656751751 40.624141275581
2.54197920777673 39.8123254040614
2.62671184803596 39.0026251711089
2.71144448829518 38.1944051807499
2.7961771285544 37.3871519922229
2.88090976881363 36.5803060098699
2.96564240907285 35.7731933894106
3.05037504933208 34.9654558582772
3.1351076895913 34.1568460779928
3.21984032985053 33.3478100550704
3.30457297010975 32.5397013074077
3.38930561036897 31.7344642309291
3.4740382506282 30.934764681895
3.55877089088742 30.1439957772963
3.64350353114665 29.365769223464
3.72823617140587 28.6038817459365
3.8129688116651 27.8622740568849
3.89770145192432 27.1449123480643
3.98243409218354 26.4558617043346
4.06716673244277 25.7992403532301
4.15189937270199 25.1791808472716
};
\addplot [semithick, black, forget plot]
table {%
0 83.9679680877326
0.221182611182912 83.0764951107084
0.442365222365823 82.4374132722446
0.663547833548735 81.9965096608266
0.884730444731647 81.699571364992
1.10591305591456 81.4923854732487
1.32709566709747 81.3207390741046
1.54827827828038 81.1304192560677
1.76946088946329 80.8672131076589
1.99064350064621 80.4769077173804
2.21182611182912 79.9052901737398
2.43300872301203 79.1203318195356
2.65419133419494 78.1371446694816
2.87537394537785 76.9723849908976
3.09655655656076 75.642709051056
3.31773916774368 74.1647731172537
3.53892177892659 72.5552334567881
3.7601043901095 70.8307463369562
3.98128700129241 69.0079680250671
4.20246961247532 67.1035547884038
4.42365222365823 65.1341628942636
4.64483483484115 63.118932639905
4.86601744602406 61.0878867551623
5.08720005720697 59.0723242414637
5.30838266838988 57.1035441002415
5.52956527957279 55.2128453329278
5.75074789075571 53.4315269409549
5.97193050193862 51.7908879257346
6.19311311312153 50.3222272887156
6.41429572430444 49.05684403133
6.63547833548735 48.0260371550102
6.85666094667026 47.2556682766956
7.07784355785318 46.7144446650517
7.29902616903609 46.3558040929703
7.520208780219 46.1331843333098
7.74139139140191 46.0000231589286
7.96257400258482 45.9097583426852
8.18375661376773 45.8158276574416
8.40493922495065 45.6716688760589
8.62612183613356 45.4307197713955
8.84730444731647 45.0464181163098
9.06848705849938 44.4737421337378
9.28966966968229 43.7398726270452
9.5108522808652 42.9114618239136
9.73203489204812 42.05516195205
9.95321750323103 41.2376252391615
10.1744001144139 40.5255039129554
10.3955827255969 39.9854502011117
10.6167653367798 39.6841163313558
10.8379479479627 39.6881545313949
};
\addplot [semithick, gray, forget plot]
table {%
0 85.1598910124061
0.217750632187712 84.7404042535428
0.435501264375424 84.2823160077801
0.653251896563136 83.787034902957
0.871002528750847 83.2562424328428
1.08875316093856 82.6916250003438
1.30650379312627 82.0947540745265
1.52425442531398 81.4662671778825
1.74200505750169 80.8057225698674
1.95975568968941 80.111006757705
2.17750632187712 79.3781082693925
2.39525695406483 78.6063519165739
2.61300758625254 77.7977771176536
2.83075821844025 76.9547445351462
3.04850885062797 76.0795830749743
3.26625948281568 75.174749428254
3.48401011500339 74.243007036295
3.7017607471911 73.2871685098434
3.91951137937881 72.3102142481089
4.13726201156653 71.3158552192073
4.35501264375424 70.308857306285
4.57276327594195 69.2923411548576
4.79051390812966 68.2682963489811
5.00826454031737 67.2382688342597
5.22601517250508 66.2033848774797
5.4437658046928 65.1644532949809
5.66151643688051 64.1223927973016
5.87926706906822 63.0779620208063
6.09701770125593 62.0313073600158
6.31476833344364 60.982096324938
6.53251896563136 59.9289205289802
6.75026959781907 58.8711215483782
6.96802023000678 57.8077090858079
7.18577086219449 56.7378438469496
7.4035214943822 55.6606983391939
7.62127212656991 54.575248713816
7.83902275875763 53.4807911853981
8.05677339094534 52.3768293409562
8.27452402313305 51.263893313558
8.49227465532076 50.1441727728153
8.71002528750847 49.0200045679494
8.92777591969619 47.893854834738
9.1455265518839 46.7703940493289
9.36327718407161 45.6548251639982
9.58102781625932 44.5535721101406
9.79877844844703 43.4735345426955
10.0165290806347 42.4217591045718
10.2342797128225 41.4058911731227
10.4520303450102 40.4338304642995
10.6697809771979 39.5135106841638
};
\draw (axis cs:3.45999999999185,9.27300786497733) node[
  anchor=north west,
  text=black,
  rotate=0.0
]{1};
\draw (axis cs:4.53999999999905,45.6360598477858) node[
  anchor=north west,
  text=black,
  rotate=0.0
]{2};
\draw (axis cs:5.95000000001164,15.7318313754852) node[
  anchor=north west,
  text=black,
  rotate=0.0
]{3};
\draw (axis cs:4.35000000000582,23.6174800061709) node[
  anchor=north west,
  text=black,
  rotate=0.0
]{4};
\draw (axis cs:10.8379479479627,39.6881545313949) node[
  anchor=north west,
  text=black,
  rotate=0.0
]{5};
\end{axis}

\end{tikzpicture}

%% file: figs/scores_1.tikz
\begin{tikzpicture}

\definecolor{color0}{rgb}{0.12156862745098,0.466666666666667,0.705882352941177}

\begin{axis}[
height=\figureheight,
legend cell align={left},
legend style={
  fill opacity=0.8,
  draw opacity=1,
  text opacity=1,
  at={(0.97,0.5)},
  anchor=east,
  draw=white!80!black
},
legend style={nodes={scale=0.7, transform shape}},
scaled y ticks=false,
tick align=outside,
tick pos=left,
width=\figurewidth,
x grid style={white!69.0196078431373!black},
xmin=-0.3, xmax=6.3,
xtick style={color=black},
xtick={0,1,2,3,4,5,6},
xticklabel style={align=center},
xticklabels={
  Training\\data,
  $\dimension=2$,
  $\dimension=3$,
  $\dimension=4$,
  $\dimension=5$,
  $\dimension=6$,
  $\dimension=7$
},
y grid style={white!69.0196078431373!black},
ylabel={Metric},
ymin=-0.00286934722698969, ymax=1.18900558990714,
ytick style={color=black},
yticklabel style={/pgf/number format/fixed,/pgf/number format/precision=3}
]
\addplot [semithick, black, mark=*, mark size=3, mark options={solid}, only marks]
table {%
0 0.967087228434522
1 1.13482945640104
2 0.966114489122617
3 0.843459260721517
4 0.889905827155426
5 0.922254112477817
6 0.951946528159182
};
\addlegendentry{$\proposedmetric{1}{\scenariosetgenerated}{\scenariosettest}{\scenarioset}$}
\addplot [semithick, color0, mark=square*, mark size=3, mark options={solid,fill=white,draw=black}, only marks]
table {%
0 0.823185627161372
1 1.1244366076906
2 0.944183236734312
3 0.805805970132817
4 0.847380020925837
5 0.875436998755388
6 0.902586129300735
};
\addlegendentry{$\wassersteinemp{1}{\scenariosettest}{\scenariosetgenerated}$}
\addplot [semithick, black, mark=square*, mark size=3, mark options={solid}, only marks]
table {%
0 0.722244106196654
1 0.0513067862791069
2 0.112577297397103
3 0.188181677682901
4 0.212667275376896
5 0.235969459567152
6 0.248280418316477
};
\addlegendentry{$\wassersteinemp{1}{\scenariosettest}{\scenariosetgenerated} - \wassersteinemp{1}{\scenarioset}{\scenariosetgenerated}$}
\end{axis}

\end{tikzpicture}

%% file: figs/scores_2.tikz
\begin{tikzpicture}

\definecolor{color0}{rgb}{0.12156862745098,0.466666666666667,0.705882352941177}

\begin{axis}[
height=\figureheight,
legend cell align={left},
legend style={
  fill opacity=0.8,
  draw opacity=1,
  text opacity=1,
  at={(0.97,0.5)},
  anchor=east,
  draw=white!80!black
},
legend style={nodes={scale=0.7, transform shape}},
scaled y ticks=false,
tick align=outside,
tick pos=left,
width=\figurewidth,
x grid style={white!69.0196078431373!black},
xmin=-0.3, xmax=6.3,
xtick style={color=black},
xtick={0,1,2,3,4,5,6},
xticklabel style={align=center},
xticklabels={
  Training\\data,
  $\dimension=2$,
  $\dimension=3$,
  $\dimension=4$,
  $\dimension=5$,
  $\dimension=6$,
  $\dimension=7$
},
y grid style={white!69.0196078431373!black},
ylabel={Metric},
ymin=0.116248559103865, ymax=1.54394438188907,
ytick style={color=black},
yticklabel style={/pgf/number format/fixed,/pgf/number format/precision=3}
]
\addplot [semithick, black, mark=*, mark size=3, mark options={solid}, only marks]
table {%
0 1.47904911721701
1 1.46809028008578
2 1.29689904561685
3 1.33389463641863
4 1.38013071105124
5 1.42432494447109
6 1.4469323381286
};
\addlegendentry{$\proposedmetric{1}{\scenariosetgenerated}{\scenariosettest}{\scenarioset}$}
\addplot [semithick, color0, mark=square*, mark size=3, mark options={solid,fill=white,draw=black}, only marks]
table {%
0 1.19900321862119
1 1.42207227611548
2 1.22261543240875
3 1.22814801710873
4 1.26541314924912
5 1.30328568272566
6 1.3259462972839
};
\addlegendentry{$\wassersteinemp{1}{\scenariosettest}{\scenariosetgenerated}$}
\addplot [semithick, black, mark=square*, mark size=3, mark options={solid}, only marks]
table {%
0 1.12392020654171
1 0.18114382377592
2 0.301048267595309
3 0.422294697229405
4 0.458330253305092
5 0.480935226148114
6 0.480419429947903
};
\addlegendentry{$\wassersteinemp{1}{\scenariosettest}{\scenariosetgenerated} - \wassersteinemp{1}{\scenarioset}{\scenariosetgenerated}$}
\end{axis}

\end{tikzpicture}

%% file: figs/wasserstein_1.tikz
\begin{tikzpicture}

\definecolor{color0}{rgb}{0.12156862745098,0.466666666666667,0.705882352941177}
\definecolor{color1}{rgb}{1,0.498039215686275,0.0549019607843137}

\begin{axis}[
height=\figureheight,
legend cell align={left},
legend style={
  fill opacity=1,
  draw opacity=1,
  text opacity=1,
  at={(0.98,0.73)},
  anchor=south east,
  draw=white!80!black
},
legend style={nodes={scale=0.7, transform shape}},
scaled y ticks=false,
tick align=outside,
tick pos=left,
width=\figurewidth,
x grid style={white!69.0196078431373!black},
xmin=-0.3, xmax=6.3,
xtick style={color=black},
xtick={0,1,2,3,4,5,6},
xticklabel style={align=center},
xticklabels={
  Training\\data,
  $\dimension=2$,
  $\dimension=3$,
  $\dimension=4$,
  $\dimension=5$,
  $\dimension=6$,
  $\dimension=7$
},
y grid style={white!69.0196078431373!black},
ylabel={Metric},
ymin=-0.00107705574366304, ymax=1.06211401283944,
ytick style={color=black},
yticklabel style={/pgf/number format/fixed,/pgf/number format/precision=3}
]
\addplot [semithick, black, mark=*, mark size=3, mark options={solid}, only marks]
table {%
0 0.735011076132639
1 1.01378714608566
2 0.824743264275033
3 0.709442782612638
4 0.731496892423181
5 0.740670682811913
6 0.748766711055716
};
\addlegendentry{$\proposedmetric{1}{\scenariosetgeneratedfake}{\scenariosettestfakesmall}{\scenariosetfake}$}
\addplot [semithick, color0, mark=square*, mark size=3, mark options={solid,fill=white,draw=black}, only marks]
table {%
0 0.625778659309074
1 1.00175121782173
2 0.797140190134458
3 0.658428788640681
4 0.681866127667784
5 0.692892414448535
6 0.704729486723327
};
\addlegendentry{$\wassersteinemp{1}{\scenariosettestfakesmall}{\scenariosetgeneratedfake}$}
\addplot [semithick, black, mark=square*, mark size=3, mark options={solid}, only marks]
table {%
0 0.546760453077211
1 0.0472498110101144
2 0.141487837124907
3 0.252537948153108
4 0.237505004205226
5 0.231416661406376
6 0.225630977508502
};
\addlegendentry{$\wassersteinemp{1}{\scenariosettestfakesmall}{\scenariosetgeneratedfake} - \wassersteinemp{1}{\scenariosetfake}{\scenariosetgeneratedfake}$}
\addplot [semithick, color1, mark=*, mark size=3, mark options={solid,fill=white,draw=black}, only marks]
table {%
0 0.458469375859693
1 0.97160814224913
2 0.674791274698836
3 0.393388697080911
4 0.426509385272418
5 0.443830361657674
6 0.456169831831441
};
\addlegendentry{$\wassersteinemp{1}{\scenariosettestfakelarge}{\scenariosetgeneratedfake}$}
\end{axis}

\end{tikzpicture}

%% file: figs/wasserstein_2.tikz
\begin{tikzpicture}

\definecolor{color0}{rgb}{0.12156862745098,0.466666666666667,0.705882352941177}
\definecolor{color1}{rgb}{1,0.498039215686275,0.0549019607843137}

\begin{axis}[
height=\figureheight,
legend cell align={left},
legend style={
  fill opacity=0.8,
  draw opacity=1,
  text opacity=1,
  at={(0.97,0.5)},
  anchor=east,
  draw=white!80!black
},
legend style={nodes={scale=0.7, transform shape}},
scaled y ticks=false,
tick align=outside,
tick pos=left,
width=\figurewidth,
x grid style={white!69.0196078431373!black},
xmin=-0.3, xmax=6.3,
xtick style={color=black},
xtick={0,1,2,3,4,5,6},
xticklabel style={align=center},
xticklabels={
  Training\\data,
  $\dimension=2$,
  $\dimension=3$,
  $\dimension=4$,
  $\dimension=5$,
  $\dimension=6$,
  $\dimension=7$
},
y grid style={white!69.0196078431373!black},
ylabel={Metric},
ymin=0.157683370913033, ymax=1.21591965291279,
ytick style={color=black},
yticklabel style={/pgf/number format/fixed,/pgf/number format/precision=3}
]
\addplot [semithick, black, mark=*, mark size=3, mark options={solid}, only marks]
table {%
0 1.05693797610373
1 1.16781800373098
2 1.01735941073578
3 1.03269180976101
4 1.03938192659559
5 1.04945617037962
6 1.0547788208027
};
\addlegendentry{$\proposedmetric{1}{\scenariosetgeneratedfake}{\scenariosettestfakesmall}{\scenariosetfake}$}
\addplot [semithick, color0, mark=square*, mark size=3, mark options={solid,fill=white,draw=black}, only marks]
table {%
0 0.856476246528326
1 1.10497297171319
2 0.908452704532271
3 0.923747539788241
4 0.934266249988523
5 0.941809162997264
6 0.952582775641875
};
\addlegendentry{$\wassersteinemp{1}{\scenariosettestfakesmall}{\scenariosetgeneratedfake}$}
\addplot [semithick, black, mark=square*, mark size=3, mark options={solid}, only marks]
table {%
0 0.802093383652447
1 0.20578502009484
2 0.442886582220572
3 0.43150888984317
4 0.424805458612267
5 0.418777331707885
6 0.409131601777954
};
\addlegendentry{$\wassersteinemp{1}{\scenariosettestfakesmall}{\scenariosetgeneratedfake} - \wassersteinemp{1}{\scenariosetfake}{\scenariosetgeneratedfake}$}
\addplot [semithick, color1, mark=*, mark size=3, mark options={solid,fill=white,draw=black}, only marks]
table {%
0 0.556141015426214
1 0.955001256281479
2 0.418652123497145
3 0.433033846220258
4 0.444387802034966
5 0.463943852866859
6 0.470256757997175
};
\addlegendentry{$\wassersteinemp{1}{\scenariosettestfakelarge}{\scenariosetgeneratedfake}$}
\end{axis}

\end{tikzpicture}

%% file: figs/correlation.tikz
\begin{tikzpicture}

\begin{axis}[
height=\figureheight,
legend cell align={left},
legend style={fill opacity=0.8, draw opacity=1, text opacity=1, draw=white!80!black},
legend style={nodes={scale=0.7, transform shape}},
scaled y ticks=false,
tick align=outside,
tick pos=left,
width=\figurewidth,
x grid style={white!69.0196078431373!black},
xlabel={$\penaltyweight$},
xmin=0, xmax=1,
xtick style={color=black},
xticklabel style={align=center},
y grid style={white!69.0196078431373!black},
ylabel={Correlation},
ymin=-0.072078872372158, ymax=1.04307281074087,
ytick style={color=black},
yticklabel style={/pgf/number format/fixed,/pgf/number format/precision=3}
]
\addplot [semithick, black]
table {%
0 0.974064672149065
0.01 0.975516904616512
0.02 0.976919063100189
0.03 0.97827899127781
0.04 0.97959834134981
0.05 0.980870462018405
0.06 0.982085814206128
0.07 0.983250618898348
0.08 0.984368551941187
0.09 0.985449397208651
0.1 0.986469838300545
0.11 0.987420369062014
0.12 0.988299398380324
0.13 0.98911228974682
0.14 0.989850672940256
0.15 0.990506265889972
0.16 0.991073886524306
0.17 0.991548137868419
0.18 0.991921775689195
0.19 0.9921842415051
0.2 0.99234234789974
0.21 0.992384097872096
0.22 0.992304819775188
0.23 0.992092015834956
0.24 0.991740114495378
0.25 0.991242254632074
0.26 0.990591415064594
0.27 0.989780426263899
0.28 0.988804378915076
0.29 0.987651975805966
0.3 0.986316762240325
0.31 0.984794536061537
0.32 0.983091165505702
0.33 0.981182627803371
0.34 0.979060877784218
0.35 0.9767175433616
0.36 0.974145601685044
0.37 0.971336198076477
0.38 0.968281260740868
0.39 0.964973302751458
0.4 0.961405040266939
0.41 0.957569437232043
0.42 0.953459751861031
0.43 0.949069584599823
0.44 0.944392224078369
0.45 0.939421872402967
0.46 0.934154014486057
0.47 0.928586870903811
0.48 0.922714949817942
0.49 0.916533834997167
0.5 0.910040994599905
0.51 0.903200699240905
0.52 0.896030588261868
0.53 0.888542260817021
0.54 0.880687421297967
0.55 0.872492270800406
0.56 0.863961560812309
0.57 0.855039119458019
0.58 0.84579426766218
0.59 0.836231985875737
0.6 0.82635826216529
0.61 0.816180081378383
0.62 0.805703808598607
0.63 0.794937734614308
0.64 0.783891592251644
0.65 0.772575898743359
0.66 0.761002003938409
0.67 0.74918203345808
0.68 0.737128825389123
0.69 0.72485586128213
0.7 0.712377192311938
0.71 0.699707361526466
0.72 0.686861323159405
0.73 0.673854360009783
0.74 0.660701999897326
0.75 0.647419932187377
0.76 0.634023925344037
0.77 0.620534785507539
0.78 0.606952804114971
0.79 0.59329628854851
0.8 0.579585736313796
0.81 0.56587710907445
0.82 0.552144310151676
0.83 0.538400521894358
0.84 0.524659484712806
0.85 0.51093446784729
0.86 0.497238231629183
0.87 0.483564622422476
0.88 0.46997158332743
0.89 0.456621827759136
0.9 0.443339746873458
0.91 0.430134995902179
0.92 0.417016651620753
0.93 0.403993209117924
0.94 0.391063198694891
0.95 0.37826300069426
0.96 0.365584132086807
0.97 0.352971298871546
0.98 0.340363827346552
0.99 0.327904153596211
1 0.315598372118764
};
\addlegendentry{LVD}
\addplot [semithick, black, dashed]
table {%
0 0.824383548981392
0.01 0.83073902115432
0.02 0.837233042356574
0.03 0.84403978995075
0.04 0.851120941171704
0.05 0.858305763017901
0.06 0.865584826098181
0.07 0.873043372829371
0.08 0.880539098454279
0.09 0.888114784519696
0.1 0.895867983556337
0.11 0.903457014323547
0.12 0.910969580222013
0.13 0.91838442991646
0.14 0.925666437897714
0.15 0.933055448145369
0.16 0.940238473479815
0.17 0.947156507758241
0.18 0.953756624251026
0.19 0.959955633724124
0.2 0.965717455170471
0.21 0.970973343668903
0.22 0.975642824498211
0.23 0.979625087636819
0.24 0.982836060787339
0.25 0.985239407305443
0.26 0.986699340302981
0.27 0.98718754847168
0.28 0.986807694286482
0.29 0.985268577841579
0.3 0.982496779915487
0.31 0.978424276563826
0.32 0.972993988375938
0.33 0.96616208401927
0.34 0.957900048762219
0.35 0.948196383630622
0.36 0.937057811117235
0.37 0.924509886486001
0.38 0.910729503398517
0.39 0.896110783626279
0.4 0.880398290723478
0.41 0.863685973117505
0.42 0.846068050069383
0.43 0.827652298998194
0.44 0.808550518952584
0.45 0.788876886509971
0.46 0.767475824894694
0.47 0.744516347436728
0.48 0.721081178708798
0.49 0.697300190738704
0.5 0.673298054692517
0.51 0.64919229262194
0.52 0.625007362247092
0.53 0.600916057457582
0.54 0.5770257499825
0.55 0.55341539480937
0.56 0.530153194043453
0.57 0.507799108218234
0.58 0.487939312033216
0.59 0.468555446582353
0.6 0.449641159036153
0.61 0.431241096878564
0.62 0.41336044485487
0.63 0.396000503984849
0.64 0.37761045004498
0.65 0.359477562165288
0.66 0.341889459694705
0.67 0.324841466813155
0.68 0.308326463773738
0.69 0.292335332825439
0.7 0.276857351313677
0.71 0.261880535123118
0.72 0.247391936224998
0.73 0.233377898394571
0.74 0.219824275248249
0.75 0.206716614676717
0.76 0.194040313569161
0.77 0.181780746473786
0.78 0.169955376308736
0.79 0.158549364701596
0.8 0.147516436932704
0.81 0.136842651009559
0.82 0.126514418008516
0.83 0.116518532423366
0.84 0.106842193878812
0.85 0.0974730216368117
0.86 0.0883990631261996
0.87 0.0796087975511703
0.88 0.0705055608152255
0.89 0.0615474566621511
0.9 0.052864343023231
0.91 0.0444173449516654
0.92 0.0362218554022912
0.93 0.0282714297725393
0.94 0.0205567057047866
0.95 0.0130686953226332
0.96 0.00578632830390686
0.97 -0.00130387827196457
0.98 -0.00819168552379107
0.99 -0.0148846872003169
1 -0.021390159503384
};
\addlegendentry{Cut-in}
\end{axis}

\end{tikzpicture}

%% file: secs/discussion.tex
\section{Discussion}
\label{sec:discussion}

One of the advantages of our proposed method for generating scenario parameters is that less assumptions are needed regarding the parameterization of the scenarios:
\begin{itemize}
	\item There is no assumption needed on a predetermined functional form of the time series data. 
	For example, in \iac{lvd} scenario, the speed is often assumed to follow a polynomial function \autocite{deGelder2017assessment}, a sinusoidal function, or a linear function \autocite{thal2020incorporating}.
	In case of a predetermined functional form, parameters are fitted to the functional form.
	In our case, the \ac{svd} automatically determines the optimal choice of parameterization without relying on a predetermined functional form.
	
	\item There is no assumption needed for the shape of the distribution of the parameters.
	For example, a particular distribution, such as a Gaussian distribution \autocite{gietelink2007phd} or a uniform distribution, may be assumed for which parameters are fitted.
	Alternatively, assumptions are made regarding the independence of the parameters \autocite{feng2021intelligent}.
	In our case, the \ac{kde} automatically adapts its shape to the data and also considers the dependence among the different parameters.
\end{itemize}
It should be noted, however, that if there is a reason to believe that one or more of the assumptions are valid, than alternative methods for generation scenario parameters that make use of such assumptions might perform equally or better than the presented method \autocite{siegel1957nonparametric}.
In most cases, it will be difficult to provide a proper justification of the assumptions regarding the functional form of, e.g., the vehicle speed, and the \ac{pdf} of the scenario parameters and the presented method will outperform methods relying on such assumptions.
In any case, the presented \ac{sr} metric provides an opportunity to verify the applicability of any assumptions regarding the scenario parameterization and parameter distributions.

The generated scenario parameters represent scenarios that could happen in real life and cover the same variety that is found in real-world traffic.
Most likely, the majority of these scenarios are straightforward for the \ac{av} to deal with.
To do an efficient assessment, the focus should be on scenarios that might lead to critical situations in which the probability of collision is high.
That is why so-called \emph{importance sampling} \autocite[Chapter~5.6]{rubinstein2016simulation} is often used for the assessment of \acp{av}, e.g., see \autocite{deGelder2017assessment, jesenski2020scalable, xu2018accelerated, zhao2018evaluation}.
With importance sampling, a different \ac{pdf}, $\densityis{\cdot}$, is used to sample scenario parameters, such that more emphasis is put on scenarios that might lead to critical situations. 
To get unbiased results, the result of a test with scenario parameters $\scenariopars$ is weighted by the ratio of the original probability density, $\densityestkde{\bandwidthmatrix}{\scenariopars}$, and the probability density of the \ac{pdf} used for importance sampling, $\densityis{\scenariopars}$ \autocite{rubinstein2016simulation, deGelder2017assessment, jesenski2020scalable, xu2018accelerated}.
Note that the importance sampling techniques explained in \autocite{deGelder2017assessment, jesenski2020scalable, xu2018accelerated} can be directly applied on the estimated \ac{pdf} $\densityestkde{\bandwidthmatrix}{\cdot}$ in \cref{eq:density est kde} of the reduced set of parameters.
In future work, our method for generating scenarios will be combined with importance sampling \autocite{deGelder2017assessment, jesenski2020scalable, xu2018accelerated} for an assessment of \iac{av}.

In some cases, one might want to sample from a conditional \ac{pdf}, e.g., in case of sampling the scenario parameters for the \ac{lvd} scenario such that the initial time gap equals a specified value. 
Sampling from \iac{kde} such that one or more parameters are predetermined is straightforward \autocite{holmes2007fast}.
In our case, sampling from $\densityestkde{\bandwidth}{\cdot}$ such that the time gap equals a specified value results in a linear constraint on the samples, because the reduced parameter vector $\svdvvecd{\scenarioindex}$ of \cref{eq:reduced parameter vector} results from a linear mapping of the original parameters $\scenariopars_{\scenarioindex}$ of \cref{eq:scenario parameters}.
In other words, one might want to sample $\scenarioparsreduced$ from $\densityestkde{\bandwidthmatrix}{\cdot}$ of \cref{eq:density est kde}, such that $\scenarioparsreduced$ is subject to the linear constraint
\begin{equation}
	\label{eq:linear constraint}
	\constraintmatrix\scenarioparsreduced = \constraintvector,
\end{equation}
where $\constraintmatrix$ and $\constraintvector$ are a matrix and vector, respectively.
In \autocite{degelder2021conditional}, an algorithm is provided for sampling from \iac{pdf} estimated using \ac{kde} such that the generated sample is subject to the constraint of \cref{eq:linear constraint}. 
The main idea of \autocite{degelder2021conditional} is to weight each parameter vector $\scenarioparsreduced_{\scenarioindex}$, $\scenarioindex \in \{1,\ldots,\scenariosnumberof\}$ in the \ac{kde} based on how closely the $\scenarioparsreduced_{\scenarioindex}$ matches the constraint \cref{eq:linear constraint}.

The presented case study considers a vehicle for which the full trajectory is predetermined. 
For the presented scenarios, this works well, but the full trajectory is not predetermined in scenarios where the actor's behavior depends on the behavior of the ego vehicle \autocite{althoff2017CommonRoad}.
To deal with such scenarios, one option is to use a driver behavior model (e.g., \autocite{treiber2000congested, kesting2007general}) with predefined parameters instead of describing the full trajectory.
The parameters of the driver behavior model may be part of $\extraparameters$.
The proposed method for generating scenario parameter values still applies in these kind of scenarios. 
Our ongoing research focuses on the assessment of \acp{av} using scenarios in which driver behavior models are used for vehicles that may respond to the ego vehicle's behavior.

Since \ac{kde} is used, the generated scenario parameters represent variations of the data.
Nevertheless, if the data do not contain scenarios that might lead to critical situations, such as an emergency braking maneuver or a reckless cut-in scenario, it is unlikely that such scenarios are generated, even if importance sampling \autocite{deGelder2017assessment, jesenski2020scalable, xu2018accelerated} is used. 
Therefore, when using the generated scenarios for the (safety) assessment of \acp{av}, it is important that there is enough data such that the data contain such scenarios.
Although there is no consensus yet on the required amount of data, some metrics have been proposed \autocite{wang2017much, degelder2019completeness} for determining whether enough data have been collected when using the data for the assessment of \acp{av}.

This work employs the Wasserstein metric to propose the \ac{sr} metric for evaluating the generated scenario parameters.
It is illustrated how our proposed metric could be used to determine the appropriate number of parameters ($\dimension$) and the type of distribution that is used to model the \ac{pdf} of the scenario parameters. 
Also, the bandwidth $\bandwidth$ or bandwidth matrix $\bandwidthmatrix$ could also be determined by optimizing the proposed metric. 
In case of the bandwidth estimation, the disadvantage is that it would require more computational resources compared to, e.g., leave-one-out cross-validation.

More research is needed to determine the influences on the optimal choice for the penalty weight $\penaltyweight$.
The case study has demonstrated one way to verify whether the initial choice of $\penaltyweight$ was appropriate, but we do not yet know \emph{why} a weight of $\penaltyweight\approx 0.25$ is an appropriate choice.
The actual choice might depend on, among others, $\scenariosnumberof$, $\scenariostestnumberof$, $\scenariosgeneratednumberof$, and the shape of the underlying distribution of the scenario parameters.
Future research with a larger data set will allow us to better determine the optimal $\penaltyweight$ and how this optimal value is influenced.

Future work involves researching the use of the proposed metric in combination with alternative methods for generating scenarios for the assessment of \acp{av}.
For example, \textcite{spooner2021generation} have used \iac{gan} \autocite{goodfellow2014generative} to create pedestrian crossing scenarios. 
One of the difficulties with \acp{gan} is to know when the \ac{gan} truly replicates the underlying distribution. 
Several metrics have been proposed \autocite{borji2019pros} to evaluate the performance of \acp{gan}, among which a metric based on the Wasserstein metric that compares the generated data with test data. 
Alternatively, our proposed metric, which also considers the training data, could be considered for evaluating \acp{gan}. 
To judge the potential of our proposed metric in this application, more research is needed.

%% file: secs/conclusions.tex
\section{Conclusions}
\label{sec:conclusions}

It is essential for the deployment of \acp{av} to develop assessment methods.
Scenario-based assessment in which test cases are derived from real-world road traffic scenarios is regarded as a viable approach for assessing \acp{av}.
This work has presented a method to generate parameterized scenarios for the use in test case descriptions for the assessment of \acp{av}.
To not rely on a small set of parameters, we have used \ac{svd} to reduce the parameters.
Parameter values for the scenarios are generated by drawing samples from the estimated \ac{pdf} of the reduced set of parameters.
To deal with the unknown shape of the \ac{pdf}, it has been proposed to estimate the \ac{pdf} using \ac{kde}.
This work has also presented a novel metric, the so-called \ac{sr} metric, based on the Wasserstein metric, for evaluating whether the generated scenario parameters represent realistic scenarios while covering the same variety that is found in real-world traffic. 

A case study has illustrated the proposed method for generating scenario parameter values using scenarios with a leading vehicle that decelerates and scenarios with a vehicle that performs a cut-in.
The case study has also illustrated that the proposed metric correctly quantifies the degree to which the generated scenario parameter values represent real-world scenarios and, at the same time, cover the same variety of scenarios that is found in real life.

Future work involves applying the proposed method for more complex scenarios, e.g., scenarios that contain several different actors, to generate scenario-based test cases for the safety assessment of \acp{av}.
Additionally, it would be of interest to apply importance sampling for \ac{av} assessment in combination with the proposed method for generating scenarios.
Other future work involves investigating the use of the proposed metric in combination with alternative methods for generating scenarios for the assessment of \acp{av}.